%% file: main.tex
\documentclass[10pt,twocolumn,letterpaper]{article}

\usepackage[pagenumbers]{cvpr} 
\usepackage[accsupp]{axessibility}  

\input{preamble}

\definecolor{cvprblue}{rgb}{0.21,0.49,0.74}

\def\confName{CVPR}
\def\confYear{2026}

\title{\ours: Toward Developmentally Grounded Pretraining and Benchmarking of Vision Foundation Models}

\author{Shengao Wang\thanks{Equal contribution.}$\;$\thanks{Project lead.}, Wenqi Wang$^*$, Zecheng Wang$^*$, Max Whitton$^*$\\
Michael Wakeham, Arjun Chandra, Joey Huang, Pengyue Zhu\\
Helen Chen\thanks{Equal contribution; work done as interns at Boston University.}, David Li$^\ddag$, Jeffrey Li$^\ddag$, Shawn Li$^\ddag$, Andrew Zagula$^\ddag$, Amy Zhao$^\ddag$, Andrew Zhu$^\ddag$\\
Sayaka Nakamura$^2$, Yuki Yamamoto$^2$, Jerry Jun Yokono$^2$\\
Aaron Mueller, Bryan A. Plummer, Kate Saenko, Venkatesh Saligrama, Boqing Gong\\
{\tt\small Boston University, $^2$Sony Group Corporation, \{wsashawn,wqwang,vicwang0,maxwh,bgong\}@bu.edu}\\
\\
\small{\url{https://shawnking98.github.io/BabyVLM-v2/}}
}

\begin{document}
 \maketitle
 \input{sec/abstract}
 \input{sec/introduction}

\input{sec/related_work}

 \input{sec/framework_overview}
 \input{sec/framework_dataset}
 \input{sec/framework_model}

\input{sec/framework_benchmark}

 \input{sec/experiment}
 \input{sec/conclusion}

{
    \small
    \bibliographystyle{ieeenat_fullname}
    \bibliography{references}
}

\input{sec/X_suppl}

\end{document}

%% file: preamble.tex
\usepackage{soul}
\usepackage[T1]{fontenc}

\newcommand{\babyvlm}{BabyVLM-V1}
\newcommand{\ours}{BabyVLM-V2}
\newcommand{\ourmodel}{``baby model''}
\newcommand{\nihtoolbox}{NIH Baby Toolbox\textsuperscript{\tiny\textregistered}}

\newcommand{\aditoolbox}{\textsl{DevCV Toolbox}}
\newcommand{\ourbenchmark}{\textsl{DevCV Toolbox}}
\newcommand{\benchmarkfont}[1]{\textsl{#1}}

\newcommand{\taskLookWhileListen}{\textsl{Looking While Listening}}
\newcommand{\taskPictureVocabulary}{\textsl{Picture Vocabulary}}
\newcommand{\taskLocalization}{\textsl{Localization}}
\newcommand{\taskLeftRight}{\textsl{Left/Right}}
\newcommand{\taskSpatialDetails}{\textsl{Spatial Details}}
\newcommand{\taskVDR}{\textsl{Visual Delayed Response}}
\newcommand{\taskMemory}{\textsl{Memory}}
\newcommand{\taskWhoHasMore}{\textsl{Who Has More}}
\newcommand{\taskSubitizing}{\textsl{Subitizing}}
\newcommand{\taskCounting}{\textsl{Object Counting}}

\newcommand{\boqing}[1]{{\color{blue}Boqing: #1}}

\newcommand{\eat}[1]{}

\usepackage{graphicx}
\usepackage{subcaption}
\usepackage{tabularx}

\usepackage{pifont}
\usepackage{booktabs}
\usepackage{xcolor}
\usepackage{colortbl}
\usepackage{multirow}
\usepackage{makecell}
\usepackage{multirow}
\usepackage{subcaption}
\definecolor{groupA}{RGB}{248,250,253}  
\definecolor{groupB}{RGB}{252,249,245}  
\definecolor{groupC}{RGB}{247,250,247}  
\definecolor{groupD}{RGB}{250,247,250}  
\definecolor{groupE}{RGB}{240,240,240}  
\newcommand{\cmark}{\textcolor{green!60!black}{\ding{51}}}

\newcommand{\xmark}{\textcolor{red!70!black}{\ding{55}}}

\usepackage{hyperref}
\usepackage[capitalize]{cleveref}

%% file: sec/abstract.tex
\begin{abstract}
Early children's developmental trajectories set up a natural goal for sample-efficient pretraining of vision foundation models. We introduce \ours, a developmentally grounded framework for infant-inspired vision-language modeling that extensively improves upon \babyvlm\ through a longitudinal, multifaceted pretraining set, a versatile model, and, most importantly, \ourbenchmark\ for cognitive evaluation. The pretraining set maximizes coverage while minimizing curation of a longitudinal, infant-centric audiovisual corpus, yielding video-utterance, image-utterance, and multi-turn conversational data that mirror infant experiences. \ourbenchmark\ adapts all vision-related measures of the recently released \nihtoolbox\ into a benchmark suite of ten multimodal tasks, covering spatial reasoning, memory, and vocabulary understanding aligned with early children's capabilities. Experimental results show that a compact model pretrained from scratch can achieve competitive performance on \ourbenchmark, outperforming GPT-4o on some tasks. We hope the principled, unified \ours\ framework will accelerate research in developmentally plausible pretraining of vision foundation models. 
 
\end{abstract}

%% file: sec/introduction.tex
\section{Introduction}
\begin{figure}[t]
    \centering
    \includegraphics[width=1\linewidth]{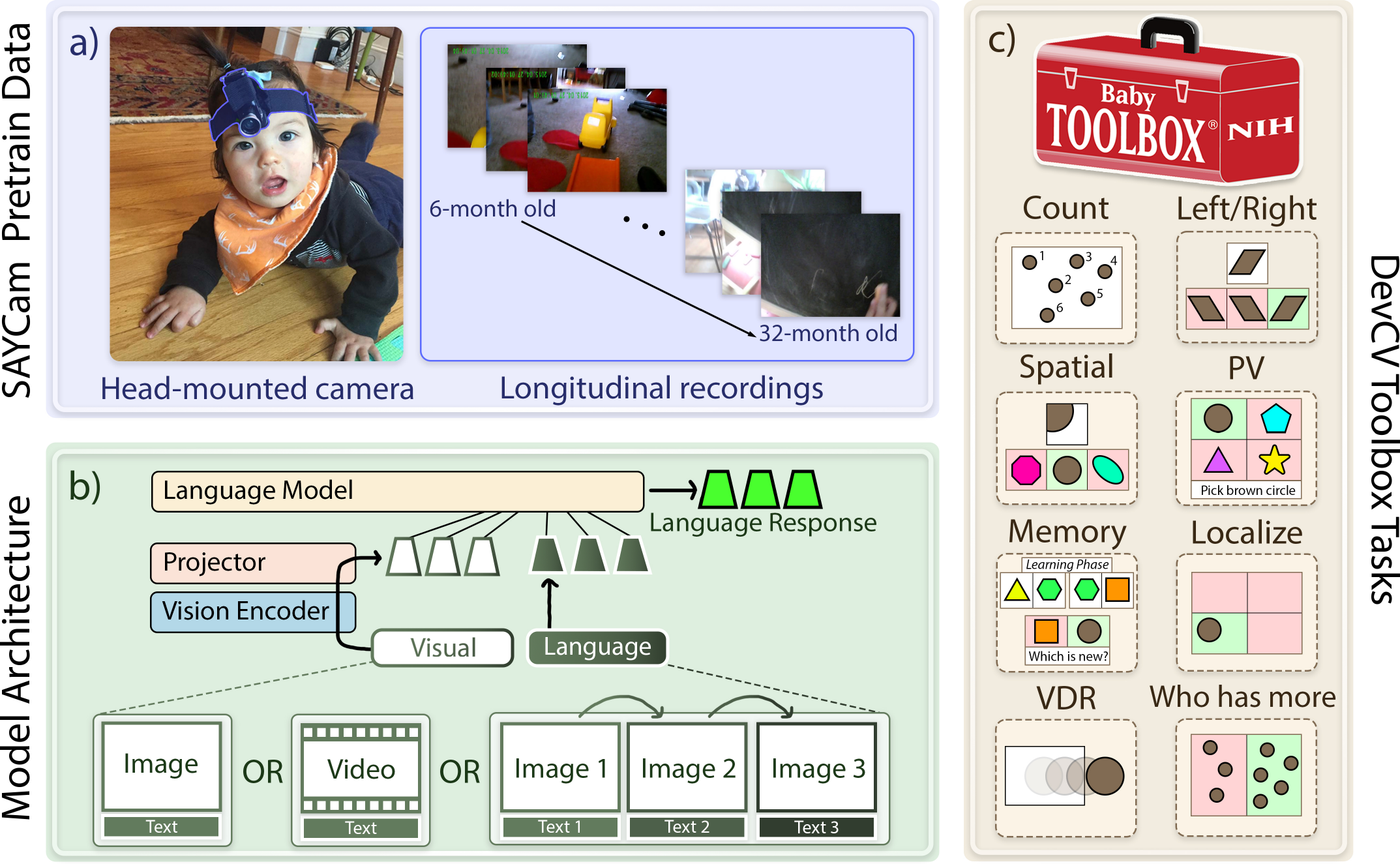}
    \vspace{-16pt}
    \caption{\textbf{\ours:} An extensive, versatile, and developmentally plausible framework for research in vision foundation models. Its (a) pretraining set  is diverse in format (video, image-utterance, and multiple turns), enabling (b) a flexible model. Its (c) benchmark  developmentally aligns with the pretraining set's age span by grounding on the newly released \nihtoolbox.}
    \vspace{-10pt}
    \label{fig:teaser}
\end{figure}

We formalize our objective: Given a longitudinal, infant-centric audiovisual sample of early children's sensory experiences
(\eg, Figure~\ref{fig:teaser}a), can we learn a foundation model (FM) that is as versatile and capable as the early children's perception? As a further challenge, can we leverage principles of developmental psychology to create a benchmark as an initial step toward artificial developmental intelligence (ADI), in both \textit{what} it is and \textit{how} to achieve it within the constraints of early children's limited sensory intake? We consider a resultant model and benchmark \textit{developmentally plausible} if the training data and desired model performance closely mirror those of early children. 

We envision that our answer to this objective, \ours, will have a threefold impact. First, by making the limited training data accessible to independent researchers and friendly to university resources, we will broaden research engagement in developing FMs~\cite{warstadt_findings_2023,hu_findings_2024} in a time when the scaling law~\cite{kaplan_scaling_2020} causes research on FMs to be  dominated by industry.
Second, we envision that ADI could advance studies in cognitive science and psychology by allowing scientists to read into early children's minds in an unprecedented way. Lastly, we believe that the broadened engagement in FMs will improve public understanding, trust, and safe use of FMs and AI in general.

Previously, Wang \etal\ proposed \babyvlm~\cite{wang_babyvlm_2025}, a scaffold for studying ADI from the lens of vision-language models (VLMs). It consists of 1) an image-text pretraining set extracted from SAYCam's
head-mounted camera recordings from three children for approximately two hours per week  from age 6 to 32 months~\cite{sullivan_saycam_2021}, 2) four intuitive and developmentally inspired benchmark tasks, and 3) a public codebase for pretraining and evaluation. \babyvlm\ pretrained a baseline VLM from scratch, whose performance, unfortunately, fell far behind the remarkable capabilities of early children~\cite{malaviya_can_2022,diesendruck_how_nodate}. Similarly, Vong \etal~\cite{vong_grounded_2024} trained a CLIP-style~\cite{radford_learning_2021} contrastive model using SAYCam, but with a narrower focus on word-referent mappings rather than general perception.
More related work is in Section~\ref{sec:related}. 

While \babyvlm\ sets up a basic framework, it lacks crucial elements. Its pretraining set only leverages about a third of SAYCam's recordings, causing it to cover only a tiny portion of the total visual intake time of a three-year-old since birth~\cite{noauthor_how_2022}. It does not support instruction tuning~\cite{zhang_instruction_2025}, which is crucial for a pretrained model to articulate its capabilities to user instructions. Importantly, its evaluation benchmarks 
are not based on any established psychology tests. Finally, the models trained in \babyvlm\ have near-zero open-set performance, and one has to postprocess their logits for evaluation. 

\input{author-kit-CVPR2026-v1-latex-/tables/V1vsV2}

This work extends \babyvlm\ to a comprehensive, extensive, and developmentally plausible framework, \ours\ (see Figure~\ref{fig:teaser}), for studying the objective posed at the beginning of the paper. Table~\ref{tab:babyvlm_comparison} contrasts the two frameworks in pretraining, instruction tuning, benchmarks, and baseline models. Notably, we provide the \textsl{Developmental Computer Vision Toolbox} (\aditoolbox)\ (see Figure \ref{fig:benchmark-tasks-figure-placeholder}), a benchmark of ten tasks designed using the \nihtoolbox~\cite{gershon_nih_2024,han_nih_2025}, which was publicly released in February 2025 as a ``universal assessment for developmental and pediatric communities''. We make minimal changes while adapting all of its vision-related measures to \ourbenchmark\ in order to maintain developmental fidelity. 

Interestingly, the \ourbenchmark\ tasks are naturally diverse in format, desiring FMs to understand individual videos and images, reason across multiple images, and solve a task in multiple turns. To account for these requirements in the pretraining data, we compile video, image-utterance, and multi-turn data from the longitudinal, infant-centric videos in SAYCam~\cite{sullivan_saycam_2021}. As in \babyvlm, we include a minimal curation process to bring our pretraining data as close to the children's sensory intake as possible. 

We validate \ours\ through extensive experiments and human performance surveys. A model trained from scratch within our \ours\ framework outperforms GPT-4o in math tasks, highlighting the potential of developmentally grounded pretraining.

%% file: author-kit-CVPR2026-v1-latex-/tables/V1vsV2.tex
\begin{table}
\centering
\small
\caption{{\ours\ extensively extends \babyvlm~\cite{wang_babyvlm_2025}.}}
\label{tab:babyvlm_comparison}
\vspace{-10pt}
    \begin{tabular}{p{0.2\linewidth} p{0.28\linewidth} p{0.37\linewidth}}
\toprule
 & {\babyvlm} & {\ours\ (Ours)} \\
\midrule
Pretraining & 67k img-utterance & 768k img-utterance \\
  &  & 181k video-utterance \\
 &  & $\;\,$63k interleaved \\
\midrule
Instruction & None & 113k examples \\
\midrule
Benchmarks & 4 tasks, intuitive & 10 tasks, grounded on\\
 &  &
\nihtoolbox\\
 & Visual vocabulary, captioning & Visual vocabulary, counting, memory, attention, spatial reasoning, localization, spatiotemporal reasoning, executive function \\
\midrule 
Models & Input: text, single img & Input: text, img, multi-img, video, multi-turn  \\
& Output: logits & Output: language \\
\bottomrule
\end{tabular}
\vspace{-10pt}
\end{table}

%% file: sec/related_work.tex
\section{Related work}
\label{sec:related}

\textbf{Vision FMs} refer to general-purpose models~\cite{bommasani_opportunities_2022} often pretrained on massive visual data~\cite{schuhmann_laion-5b_2022, caron_web-scale_2024,miech_howto100m_2019,wang_internvid_2023}. They can tackle many vision tasks via a unified interface, such as CLIP~\cite{radford_learning_2021}, ALIGN~\cite{jia_scaling_2021}, BLIP~\cite {li_blip_2022,li_blip-2_2023}, SAMs~\cite{kirillov_segment_2023,ravi_sam_2024}, and vision LLMs~\cite{comanici_gemini_2025,openai_gpt-4o_2024, li_multimodal_2025,bai_qwen25-vl_2025}. The development of these powerful models hinges critically on pretraining~\cite{erhan_why_2010,bommasani_opportunities_2022,chen_vlp_2023}, a process that trains a model on a large, generic dataset before tuning it to any downstream tasks. 

\noindent\textbf{Sample-efficient pretraining.}
While FMs have been relying on the scaling law~\cite{kaplan_scaling_2020}, sample-efficient pretraining has gained momentum recently in the language~\cite{warstadt_findings_2023} and medical~\cite{sun_data-efficient_2025} domains. To the best of our knowledge, \babyvlm\ was the first of this kind in vision, and we further their effort with a more comprehensive and extensive framework.

\input{tables/benchmark-comparison}

\noindent\textbf{Cognitively plausible benchmarking.} \babyvlm~\cite{wang_babyvlm_2025} designs four developmentally inspired tasks, which unfortunately lack grounding on established psychological tests. DevBench~\cite{tan_devbench_nodate} and KIVA~\cite{yiu_kiva_2024} draw inspiration from kid-oriented tests, yet they are more age-advanced than our pretraining data. Other cognitively plausible benchmarks have a narrower focus, such as Zorro~\cite{huebner_babyberta_2021}, LRS~\cite{kosoy_comparing_2023}, InfLevel~\cite{weihs_benchmarking_2022}, CoreCognition~\cite{li_core_2025}, and MEWL~\cite{jiang_mewl_2023}, and ModelVsBaby~\cite{sheybani_modelvsbaby_2024}. Table~\ref{tab:related_benchmark} summarizes the differences. 

\noindent\textbf{Tools assessing neurodevelopment in children.} Our benchmark tasks are grounded on the \nihtoolbox~\cite{gershon_nih_2024}, a standardized tool released in February 2025 for assessing neurodevelopment in children. It is not only more recent but also more comprehensive and normed than alternatives, such as the Bayley Scales Of Infant and Toddler Development~\cite{balasundaram_bayley_2025}, Mullen Scales of Early Learning~\cite{dumont_mullen_2014}, and Battelle Developmental Inventory~\cite{newborg_battelle_2005}. Besides, its design for clinical use validates its credibility over the psychological tests used in research settings.

%% file: tables/benchmark-comparison.tex
\begin{table*}
\centering
\caption{{Comparison of existing developmentally inspired benchmarks.}
}
\label{tab:related_benchmark}
\vspace{-10pt}
\resizebox{\textwidth}{!}{
\begin{tabular}{lcccccccccc}
\toprule
\textbf{Benchmark} &
\textbf{Developmental} &
\textbf{Task Diversity} &
\textbf{Multimodal} &
\textbf{Train} &
\textbf{Val} &
\textbf{Test} &
\textbf{In-Domain} &
\textbf{OOD} &
\textbf{Human Data} &
\textbf{Model} \\
\midrule
DevBench \cite{tan_devbench_nodate}       & \cmark & \cmark & \cmark & \xmark & \xmark & \cmark & \xmark & \cmark & \cmark & \xmark \\
Labeled-S \cite{vong_grounded_2024}      & \cmark & \xmark & \cmark & \cmark & \xmark & \cmark & \cmark & \xmark & \xmark & \xmark \\
ModelVsBaby \cite{sheybani_modelvsbaby_2024}    & \cmark & \xmark & \cmark & \cmark & \cmark & \cmark & \xmark & \cmark & \cmark & \xmark \\
MEWL \cite{jiang_mewl_2023}           & \xmark & \cmark & \cmark & \cmark & \cmark & \cmark & \cmark & \xmark & \cmark & \xmark \\
Zorro  \cite{huebner_babyberta_2021}         & \cmark & \xmark & \xmark & \cmark & \xmark & \cmark & \cmark & \xmark & \xmark & \cmark \\
InfLevel  \cite{weihs_benchmarking_2022}      & \cmark & \xmark & \cmark & \xmark & \xmark & \cmark & \xmark & \cmark & \cmark & \xmark \\
LRS \cite{kosoy_comparing_2023}            & \cmark & \cmark & \xmark & \xmark & \xmark & \cmark & \xmark & \cmark & \xmark & \xmark \\
CoreCognition \cite{li_core_2025}  & \cmark & \cmark & \cmark & \xmark & \xmark & \cmark & \xmark & \cmark & \cmark & \xmark \\
BabyVLM \cite{wang_babyvlm_2025}   & \xmark & \cmark & \cmark & \cmark & \xmark & \cmark & \cmark & \xmark & \xmark & \cmark  \\
\midrule
\textbf{\ourbenchmark\;(Ours)} & \cmark & \cmark & \cmark & \cmark & \cmark & \cmark & \cmark & \cmark & \cmark & \cmark \\
\bottomrule
\end{tabular}
}
\vspace{-5pt}
\end{table*}

%% file: sec/framework_overview.tex
\section{\ours}
\label{sec:framework-overview}

%% file: sec/framework_dataset.tex
\begin{figure*}
    \centering
    \small
    \begin{tabular}{c}
         \includegraphics[width=\linewidth]{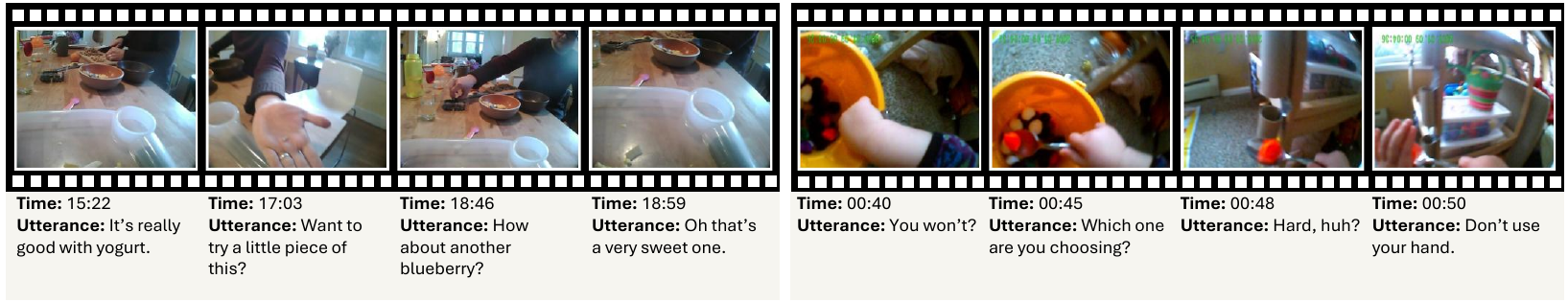} 
         \vspace{-10pt}
         \\
     \includegraphics[width=\linewidth]{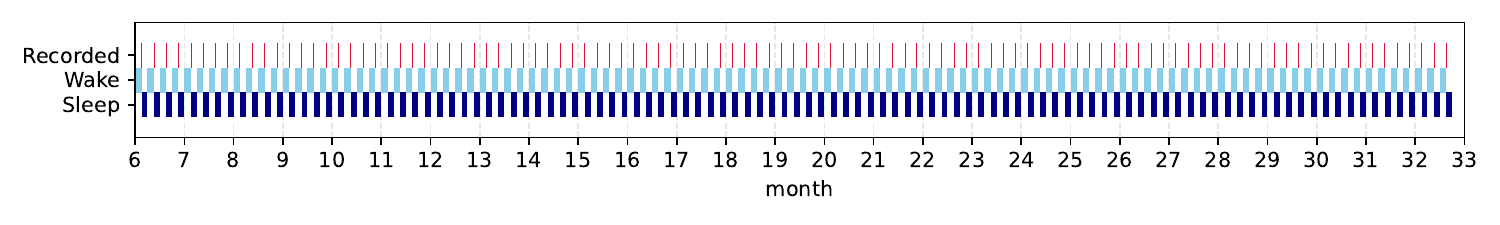}
    \end{tabular}
    \vspace{-20pt}
    \caption{Top: Video frames and utterances recorded from the infants' view. Bottom: Recorded wake time \vs\ wake/sleep time for the ages of 6 months to 32 months in SAYCam~\cite{sullivan_saycam_2021}. }
    \label{fig:saycam-example-time-distribution}     
    \vspace{-10pt}
\end{figure*}

\subsection{Data source \& the pretraining set} \label{sec:training-set}
We describe SAYCam, the developmental data source, followed by our minimal process to curate the pretraining set.

\noindent\textbf{SAYCam~\cite{sullivan_saycam_2021}:}
The developmental plausibility of our work hinges on the use of a visual-audio-text corpus that faithfully samples what early children have seen and heard
by a certain age, which requires the corpus to be 1) longitudinal and 2) infant-centric. To accomplish this, we use the SAYCam dataset~\cite{sullivan_saycam_2021}, which is accessible to all nonprofit institutes, {and will include BabyView}~\cite{long_babyview_2025} in future work. 
SAYCam contains egocentric recordings from three infants (left of Figure~\ref{fig:teaser}a) taken once every week from roughly 6 to 32 months old. Each recording is approximately two hours, and the recordings total 478 hours (see bottom of Figure~\ref{fig:saycam-example-time-distribution} for the recorded time \vs\ wake and sleep time~\cite{noauthor_how_2022}). Notably, the utterances found in SAYCam are mostly from caregivers providing simple verbal instructions and descriptions to the infants (top of Figure~\ref{fig:saycam-example-time-distribution}). BabyView~\cite{long_babyview_2025} is an ongoing effort in the same spirit as SAYCam, but at a larger scale and with extra gyroscope/accelerometer sensors. 

\noindent\textbf{Data split \& the pretraining set.}
To maximize our use of the SAYCam corpus, we designate all video clips containing speech to the {pretraining} split, and evenly divide the remaining clips into {validation} and {test} splits. Their relative sizes are approximately 3:1:1, respectively. We then apply minimal processing to facilitate model pretraining while observing the children's sensory intake  as much as possible. Specifically, we transcribe all utterances, which are almost all from caregivers, using Azure Speech Recognition~\cite{noauthor_azure_nodate}.
We then construct three types of pretraining data.  
\begin{itemize}
    \item \textbf{Video-utterance pairs.} We segment the camera recordings into short clips based on transcript boundaries, with each clip corresponding to exactly one utterance. We then drop the video clips shorter than 0.5 seconds or with a transcript confidence score below 0.3. Further, we compute video-utterance similarities using X-CLIP~\cite{ma_x-clip_2022} and only retain the video-utterance pairs with similarities greater than $0.1$. This process leaves approximately 181k video clips in our pretraining set, a total of 138 (out of 478) hours. We pad 1 second to either side of the clips. 
    \item \textbf{Image-utterance pairs.} Following BabyVLM-V1, we sample at 1 FPS from the video-utterance pairs and compute the CLIP similarity~\cite{radford_learning_2021} between each frame and its utterance. Only frames with CLIP similarities $>$ 0.2 are retained, resulting in 768k image-utterance pairs in total.
    \item \textbf{Interleaved text and images.} We create sequences of interleaved images and utterances from consecutive video segments, aiming to enable downstream capabilities that involve conversations. For each video segment, we pair the frame that has the highest CLIP similarity with its associated utterance and use a sliding window over the resulting image-utterance pairs to construct the interleaved sequences. We randomly choose a window size between 4 and 8 and employ a stride of half the window size, resulting in 63k interleaved sequences.
\end{itemize}
Unlike \babyvlm's image-utterance pairs, the mixing of three pretraining data formats prepares models for diverse downstream tasks, which can involve videos, multiple or single images, and even multi-turn conversations. 

%% file: sec/framework_model.tex
\eat{
\begin{table}[t]
\centering
\caption{\textbf{Summary of benchmark tasks.}
Each tasks lists its training and testing splits, along with the number of choices. \boqing{Please add the age information to the table. It's available here: \url{https://babytoolbox.zendesk.com/hc/en-us/article_attachments/38301189325716}}}
\label{tab:benchmark_summary}
\vspace{-10pt}
\setlength{\tabcolsep}{6pt} 
\renewcommand{\arraystretch}{1.1} 
\begin{tabular}{lccc}
\toprule
\textbf{Benchmark} & \textbf{Train} & \textbf{Test} & \textbf{Age (months)} \\
\midrule
Counting & 13.7k & 3.0k & 25-42 \\
Memory & 10.0k & 0.5k & 22-42 \\
Visual delay response & 5.2k & 0.9k & 22-42 \\
Left/Right & 12.3k & 2.3k & 1-42 \\
Spatial Details & 11.8k & 1.2k & 1-42 \\
Localize & 12.3k & 2.1k & 1-42 \\
Picture Vocabulary & 63.9k & 1.2k & 25+ \\
Who Has More & 23.4k & 3.1k & 25-42 \\
\midrule
\textbf{Total} & 152.6k & 14.3k & -- \\
\bottomrule
\end{tabular}
\end{table}
}

\subsection{Pretraining \& fine-tuning a \ourmodel}
\label{sec:model}
Using our pretraining corpus, we pretrain a \ourmodel, which uses a language model (LLaMA-1.1B~\cite{zhang_tinyllama_2024,touvron_llama_2023}) as a versatile interface to probe various capabilities of a visual encoder (ViT-L-16~\cite{dosovitskiy_image_2021}, 300M parameters). A lightweight MLP connector~\cite{llava1.5} projects visual features into the language space. This model architecture (Figure~\ref{fig:teaser}b) is the same as \babyvlm's BabyLLaVA-Llama. We pretrain the entire model from scratch using the three-stage pipeline described in Appendix~\ref{app:model-training}. 
Finally, we fine-tune the model using a small, curated instruction set consisting of the tasks as in \ourbenchmark, which we describe next.

%% file: sec/framework_benchmark.tex


\subsection{Age-appropriate \ourbenchmark}
\label{sec:benchmark}
Our objective with \ours\ entails designing benchmark tasks that test \textit{age-appropriate} visual skills given our pretraining data's age span.
However, we acknowledge that developmental benchmarking is an ongoing and rapidly evolving field of research. Early children's growth rates vary significantly, and among psychologists and cognitive scientists, substantial conceptual and methodological disagreements exist regarding the notion of developmental intelligence and how to properly probe, measure, and benchmark it~\cite{han_nih_2025}. How can we properly define ADI, then, given the inconsistent measurement techniques in human developmental research? To  answer this, we consult with two experienced psychologists specializing in development and learning. Numerous meetings with them led us to the timely \nihtoolbox, over which we ground the design of our benchmark, 
\ourbenchmark.


\subsubsection{Background: \texorpdfstring{\nihtoolbox}{NIH Baby Toolbox®}}
In February 2025, a multi-institutional team solicited by NIH~\cite{gershon_nih_2024} released the \nihtoolbox, envisioning it as a 
standardized evaluation of neurodevelopmental intelligence in infants~\cite{han_national_nodate}. The \nihtoolbox\ divides developmental function into three domains: Cognition, Motor, and Social-Emotional, where the Cognition domain includes the subdomains of Language, Executive Function/Memory, and Math, each consisting of some number of specific tests, known in the toolbox as \textit{measures}. See Table~\ref{tab:benchmark-mapping} for a summary of these measures and Appendix~\ref{appedix:benchmarks}  for technical details.

\begin{figure*}
    \centering
    \includegraphics[width=1\linewidth]{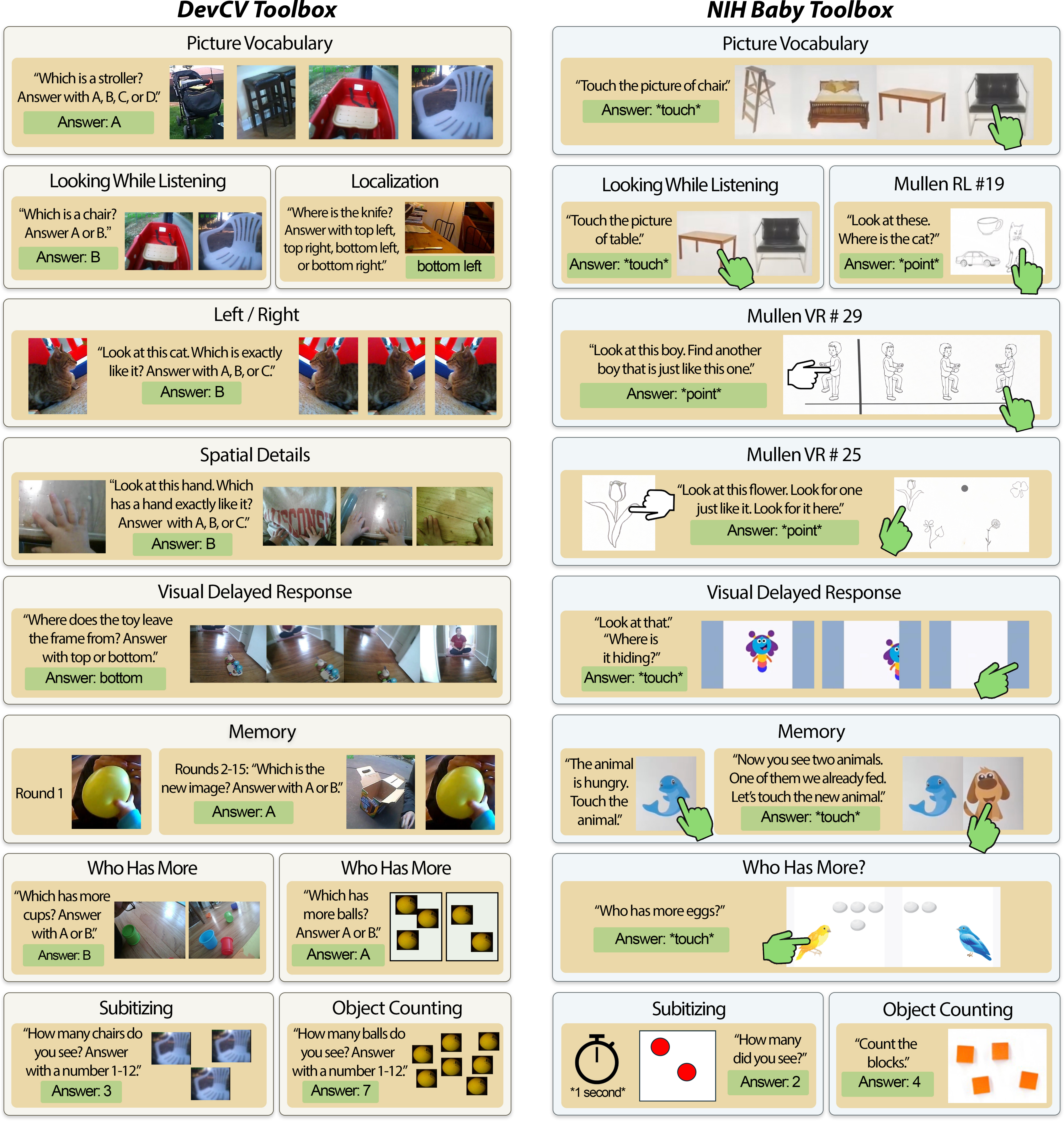}
    \vspace{-20pt}
    \caption{\aditoolbox\ tasks and their corresponding \nihtoolbox\ measures}
    \vspace{-10pt}
    \label{fig:benchmark-tasks-figure-placeholder}
\end{figure*}
\input{tables/benchmark-mapping}

\subsubsection{\textbf{\ourbenchmark}}
\label{sec:devcv_toolbox}
In this section, we develop a computer vision counterpart, called \benchmarkfont{task} for clarity, for every vision-related \textit{measure} in the \nihtoolbox,  leading to ten tasks in our \ourbenchmark, which are summarized in Table~\ref{tab:benchmark-mapping} and illustrated in Figure~\ref{fig:benchmark-tasks-figure-placeholder}. 

\noindent\textbf{The need to adapt \textit{measures} to \benchmarkfont{tasks}.} Unlike the practice in computer vision, most of the measures originally found in the \nihtoolbox\ 1) have only a couple of test examples and 2) are human-oriented but not accessible to AI models. Additionally, the cartoon stimuli in \nihtoolbox\;are out-of-domain from our pretraining set, complicating the intended ``skill benchmark'' purpose. Hence, we adapt the measures to computer vision tasks by standardizing their format and equipping each task with thousands of naturalistic examples  (see Table~\ref{tab:benchmark-mapping}), separated into instruction-tuning, validation, and test sets according to the splits defined in the pretraining stage. 

We construct the tasks using SAYCam to ensure that the examples are in the same domain as the pretraining data, thereby focusing the benchmarking on the models' in-domain cognitive capabilities. To provide an additional tool to evaluate models' generalizability, we also compile an out-of-domain test set using Ego4D~\cite{grauman_ego4d_2022} with the same techniques. Below, we detail the construction of \taskPictureVocabulary\ as a representative example, and briefly describe the rest. See Appendix~\ref{appedix:benchmarks} for more details on the construction of \ourbenchmark. 


\begin{figure*}
    \centering
    \includegraphics[width=0.9\linewidth]{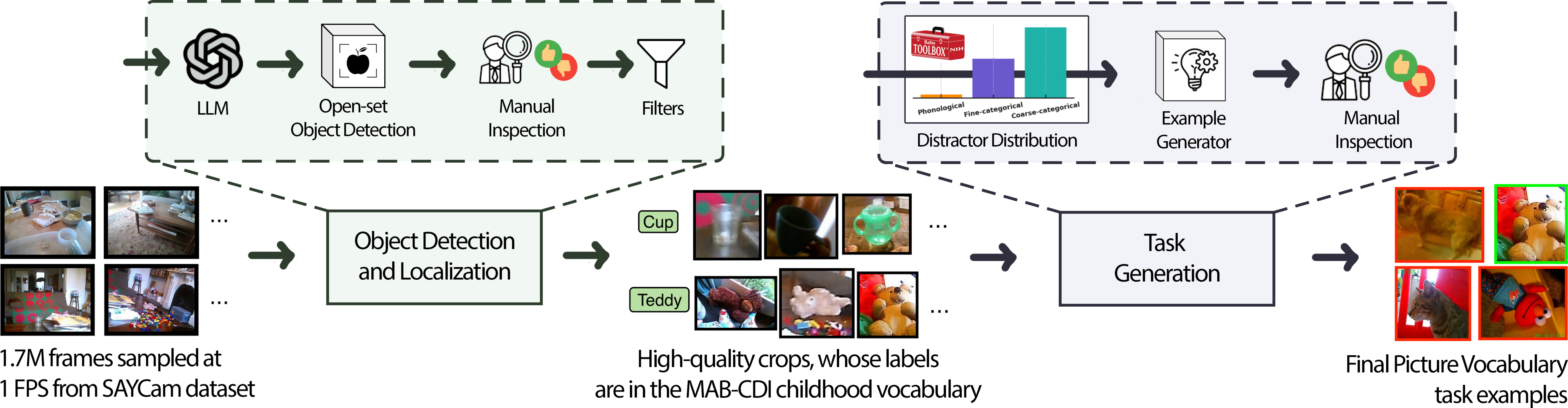}
    \vspace{-5pt}
    \caption{Pipeline to adapt the picture vocabulary measure in \nihtoolbox\ to \ourbenchmark.}
    \label{fig:pv-pipeline}
    \vspace{-5pt}
\end{figure*}

\noindent\textbf{Picture vocabulary ($\ge$ 25 months):} 
The top right of Figure~\ref{fig:benchmark-tasks-figure-placeholder} shows the original picture vocabulary (PV) measure in the \nihtoolbox, which assesses the Receptive Language of children aged 25 months and older. Participants are presented with four clipart images on an iPad, and an audio prompt instructs them to touch the named image. 

We adapt PV to \ourbenchmark\ using the pipeline in Figure~\ref{fig:pv-pipeline}, to replace the clipart in the \nihtoolbox\ with \textit{objects and actions} detected from SAYCam video frames. Concretely, we sample frames at 1~FPS, label all objects and actions present using manual transcripts and GPT-4o, and then crop out regions for each label using Grounding-DINO~\cite{liu_grounding_2024}. Low quality crops and labels beyond the child-oriented MAB-CDI vocabulary~\cite{marchman_macarthur-bates_2023} are removed. Each PV example (\eg, the top left of Figure~\ref{fig:benchmark-tasks-figure-placeholder}) consists of a language prompt, a target image corresponding to the prompt, and three distractor images, we then construct the examples in a round-robin manner for diversity. The target and distractor images are related either semantically or phonologically in \nihtoolbox; therefore, we derive a distractor distribution over phonology and semantics from the toolbox and then sample distractor images accordingly. We manually screen the process to ensure quality and diversity. Appendix~\ref{appedix:benchmarks} presents more details. 


\noindent\textbf{Other tasks.} We describe the other tasks in \ourbenchmark\ briefly. 
\label{taskdescriptions}
Construction details are in Appendix~\ref{appedix:benchmarks}. 
\begin{enumerate}
    \item \textbf{Looking while listening} (6--24 months) shows infants two clipart objects, and plays an audio prompt describing one of them. Eye tracking is used to detect the participant's response. We replace clipart with natural objects from SAYCam, and eye tracking with multiple choice. 
    

    \item \textbf{Localization} / \textbf{Mullen visual receptive language \#19} (1--42 months) tests an infant's ability to point at sketched objects as they are named. We task a model with localizing an object in a natural video frame.
    

    \item \textbf{Left/Right} / \textbf{Mullen visual reception \#29} (1--42 months) measures an infant's attention to detail by instructing them to match objects by orientation. 

    
    \item \textbf{Spatial details} / \textbf{Mullen visual reception \#25} (1--42 months) measures attention to detail in identical objects among distractors of the same type.

    
    \item \textbf{Visual delayed response} (22--42 months) shows infants a creature moving behind one of two occluders, and after a short pause, instructs them to tap the target occluder. We adapt it to \ourbenchmark\ by mining video clips with prominent objects moving out of the field of view.  



    
    
    \item \textbf{(Delayed) memory} (22--42 months) involves multiple turns, each presenting a pair of animals. Participants are asked to ``feed'' the new animal appearing for the first time, and they receive corrective feedback during the early rounds.

    
    \item \textbf{Who has more} (25--42 months) shows two images with the same shape in different quantities and asks which image has more. We replace the shape with natural objects as one sub-task, and use entire natural video frames for the other sub-task. 

    
    \item \textbf{Subitizing} (25--42 months) refers to the rapid identification of the number of items in a small set. An infant sees one to four identical shapes for one second, and then an audio prompt requests the count.

    
    \item \textbf{Object counting} (25--42 months) evaluates a child's ability to count up to 12 colored shapes on a screen.
\end{enumerate}
During evaluation, we employ accuracy as the metric. These tasks cover all cognitive measures in \nihtoolbox\ except the \textit{non-visual} MacArthur-Bates language (9--30 months, 7--18 months), familiarization (6--21 months), verbal counting (25--42 months), and verbal arithmetic (37--42 months). Adult performance data on these tasks confirms the validity of our \ourbenchmark\ (see \textit{Human performance} in Tables \ref{tab:main_result_in_domain} and Appendix~\ref{app:humansurvey} for details). In future work, we hope to complete a survey of children's performance on \ourbenchmark.


%% file: tables/benchmark-mapping.tex
\begin{table*}
\centering
\setlength{\tabcolsep}{5pt} 
\caption{{\aditoolbox\ \benchmarkfont{tasks} and their corresponding \nihtoolbox\ \textit{measures} (EF/M stands for Executive Function/Memory).}}
\label{tab:benchmark-mapping}
\vspace{-10pt}
\renewcommand{\arraystretch}{1.1} 
\begin{tabular}{lrr||lrr}
\toprule
\textbf{\aditoolbox\ \benchmarkfont{tasks}} &\textbf{ \#Instruct/Val/Test }& \textbf{\#Imgage}  & \textbf{\nihtoolbox\ \textit{measures}} & \textbf{Months}  & \textbf{Subdomain}\\
\midrule
\taskLookWhileListen & 0 / 0 / 1.2k &2 imgs  & Looking While Listening&$\;$6-24 & Language \\
\taskPictureVocabulary &22.4k / 0.7k / 1.1k&  4 imgs& Picture Vocabulary&25+ & Language\\
\taskLocalization&12.3k / 2.1k / 2.1k& 1 img& Mullen Receptive Language \#19 & 1-42  & Language\\
\hline
\taskLeftRight &12.3k / 2.2k / 2.3k &4 imgs&Mullen Visual Reception \#29& 1-42  & EF/M\\
\taskSpatialDetails &11.8k / 1.2k / 1.2k& 4 imgs &Mullen Visual Reception \#20& 1-42 & EF/M\\
\taskVDR &2.6k / 0.4k / 0.5k  &5-8 imgs& Visual Delayed Response &22-42 & EF/M\\
\taskMemory &12.0k / 0.3k / 0.3k& 29 imgs & Delayed Memory& 22-42 & EF/M\\
\hline
\taskWhoHasMore\ (synthetic) &11.2k / 1.8k / 1.8k& 2 imgs & Who Has More& 25-42  & Math\\
\taskWhoHasMore\ (naturalistic) &12.4k / 1.9k / 2.2k&2 imgs & Who Has More& 25-42  & Math\\
\taskSubitizing\ (synthetic)& 0 / 0 / 1.9k& 3 imgs & Subitizing&25-42 & Math\\
\taskSubitizing\ (naturalistic)& 0 / 0 / 0.2k& 3 imgs & Subitizing&25-42 & Math\\
\taskCounting &13.7k / 3.0k / 3.0k& 1 img & Object Counting& 25-42 & Math\\
\bottomrule
\end{tabular}

\end{table*}

%% file: sec/experiment.tex
\section{Experiments}

\eat{
\boqing{Experiment structure: }
\begin{itemize}
    \item Main table, in-domain \& out-of-domain
    \item Instruction tuning vs.\ task-specific tuning
    \item Various pretraining strategies, including a randomly initialized checkpoint (always fix the instruction tuning strategy)
    \item Instruction tuning of opensource models of various sizes vs.\ ours
    \item Overall performance vs.\ amount of instruction tuning data
    \item Mixing ratios in instruction tuning? (optional)
    \item Conversational test scenarios (optional)
\end{itemize}
\boqing{Besides the above, we also need to analyze what's challenging and what's easy for our model and for Gemini/GPT. If Gemini fails, we should try to understand why.}
}

We design experiments about the key elements of the \ours\ framework, aiming to validate the quality of the \ourbenchmark, as well as illustrate the effectiveness of our training data and training recipe. Meanwhile, the experiments position our \ourmodel\ in context across three cognitive subdomains and ten tasks. Note that we exclude two tasks, \taskSubitizing\ and \taskLookWhileListen, from the majority of the experiments to test our models' generalization on unseen tasks near the end. Implementation details are in Appendix~\ref{app:additionalexp}.

\input{author-kit-CVPR2026-v1-latex-/tables/exp_in_domain}

\subsection{Examining \textbf{\ourbenchmark}}
\textbf{Overall quality.} We validate the quality of \ourbenchmark\ by conducting human surveys, detailed in Appendix~\ref{app:humansurvey}. As shown in Table~\ref{tab:main_result_in_domain}, the human volunteers recruited in our home institute achieved near-perfect accuracy on the the executive functioning/memory subdomain (\taskSpatialDetails, \taskMemory, \taskVDR) and the math tasks of \taskCounting\ and \taskWhoHasMore. Their accuracy on \taskLocalization\ is slightly low (87.3\%), and a follow-up revealed that it could improve when the volunteers were instructed to spend more time on the task. 

\noindent\textbf{Differentiating capability.} Table~\ref{tab:main_result_in_domain} also demonstrates that, between \texttt{Human performance} and \texttt{Random guess}, there is a sufficiently big room for differentiating various models. Indeed, the proprietary GPT and Gemini models are on the upper end, while smaller open-source models and our \ourmodel\ are on the lower end, with larger open-source models such as Qwen3-VL-30B-A3B in between, indicating that the tasks in \ourbenchmark\ are challenging but solvable.

\noindent\textbf{Developmental fidelity.} \ourbenchmark\ should developmentally align with the pretraining data's age span (6--32 months). Hence, we are in the process of performing a large-scale children survey about \ourbenchmark\ using the Children Helping Science platform~\cite{noauthor_children_nodate}, though this survey will take a couple of years per our estimation.

\begin{figure}
    \centering
    \includegraphics[width=\linewidth]{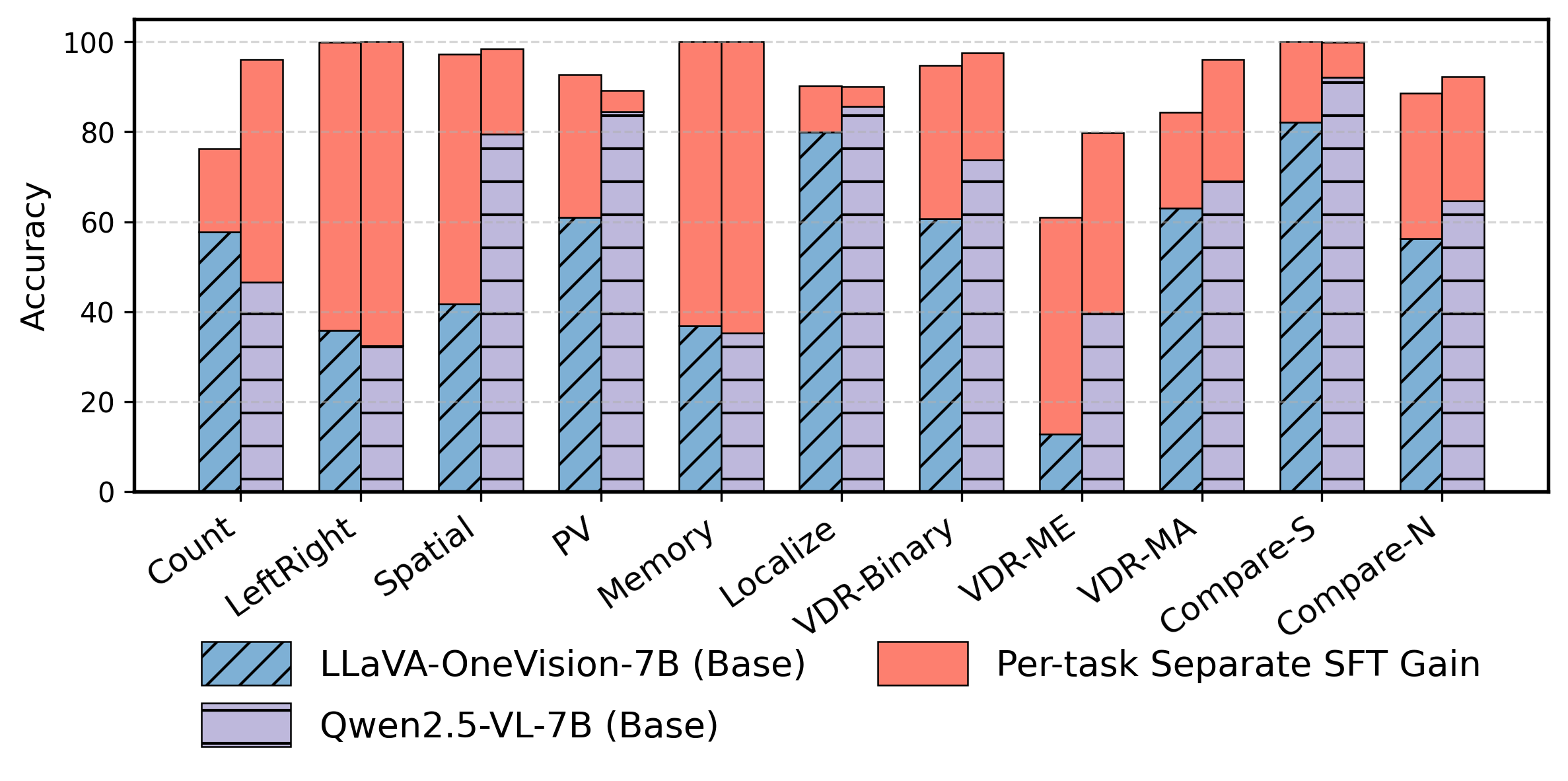}
    \vspace{-25pt}
    \caption{Task-specific supervised finetuning (SFT) of LLaVA-OneVision-7B and Qwen2.5-VL-7B.}
    \vspace{-10pt}
    \label{fig:opensource_ft}
\end{figure}

\subsection{Validating the instruction-tuning dataset}
Instruction tuning mitigates the mismatch between pretraining and downstream tasks, steering models toward target objectives. To evaluate the effectiveness of our instruction data, we supervise fine-tune three models under two strategies. As shown in Figure~\ref{fig:opensource_ft}, we fine-tune LLaVA-OneVision-7B~\cite{li2024llavaonevisioneasyvisualtask} and Qwen2.5-VL-7B~\cite{bai_qwen25-vl_2025} separately on each task (see Appendix~\ref{app:model-training} for details). Consistent and substantial gains, highlighted by the red top bars, indicate that the instruction data effectively guides models toward the downstream tasks in \ourbenchmark.
We further explore a second strategy that merges all instruction data into a single set; Appendix~\ref{app:additionalexp} provides a comparison of the two strategies on our \ourmodel. The second setting is adopted as the default for our \ourmodel.

\input{author-kit-CVPR2026-v1-latex-/tables/exp_synthetic_utterance}
\subsection{Ablating the pretraining data}
\label{sec:pretrain_data_ablation}

The speech transcripts in our pretraining set could be noisy because the naturalistic child-directed utterances are often misaligned with the children's visual intake. We study their impact on the pretrained models by replacing the transcripts with video captions generated by GPT-4o (see Appendix~\ref{app:additionalexp} for how we prompt GPT-4o). We train \ourmodel-synthetic on this altered pretraining dataset and present the results in Table~\ref{tab:synthetic_utterance}. Overall, the synthetic captions improve performance, especially on tasks that demand semantic reasoning (\taskCounting) and a long attention window (\taskMemory). However, the gains are modest, suggesting that our minimally curated pretraining set already provides strong supervision. 


\subsection{Inspecting the \ourmodel}
The overall performance of our \ourmodel\ in Table~\ref{tab:main_result_in_domain} is encouraging, on par with open-source models of comparable size, and notably competitive even against some larger open-source models on certain tasks. Of course, one could argue that those models are not fine-tuned under the \ours\ framework, but they are likely trained on much larger datasets than ours.

To further stretch the \ourmodel, we study its generalization along two axes: 1) out-of-domain generalization and 2) performance over previously {unseen tasks}.

\noindent\textbf{Out-of-domain generalization.}
We have created a sibling of \ourbenchmark\ by replacing SAYCam with Ego4D. Both are about egocentric videos, but Ego4D is from the perspective of grown-ups. The overall accuracy of the \ourmodel\ on this sibling benchmark is 41.1\% (vs.\ 31.8\% of random guess), significantly lower than its in-domain performance (55.2\%) on \ourbenchmark. We conclude that \ourmodel\ can generalize beyond its training domain to some degree, but it is far from human infants' remarkable generalization capabilities. Appendix~\ref{app:additionalexp} further tests \ourmodel on the original \nihtoolbox. 



\noindent\textbf{Unseen tasks.} We have excluded \taskLookWhileListen\ and \taskSubitizing\ from the instruction tuning, which are thus unseen during the \ourmodel\ training. While the two tasks are in spirit similar to \taskPictureVocabulary\ and \taskCounting, respectively, the \ourmodel\ yields near-random-guess results on them. We will address this issue in future work by improving the instruction tuning algorithm.

\begin{figure}
    \centering
    \includegraphics[width=\linewidth]{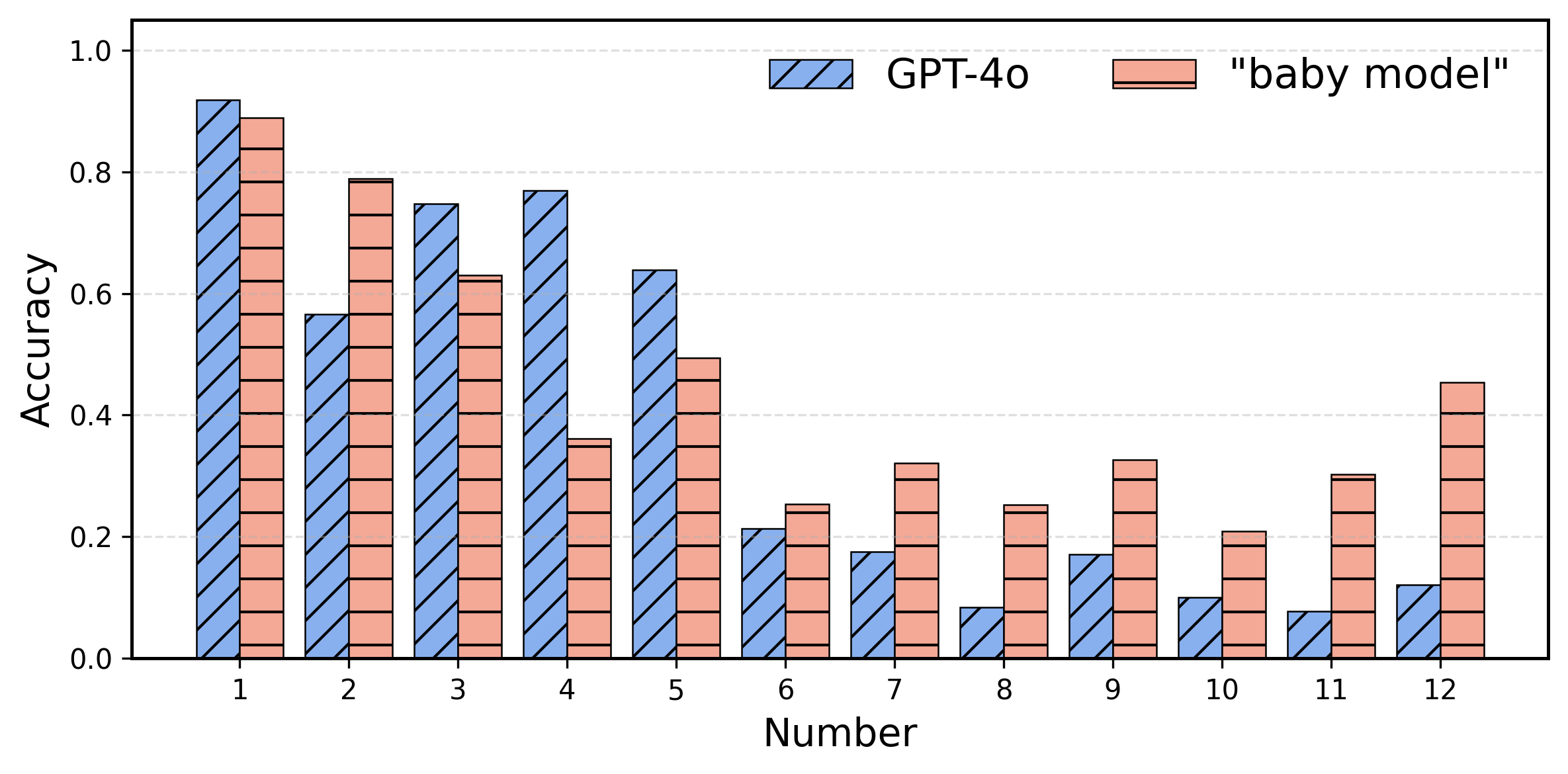}
    \vspace{-25pt}
    \caption{GPT-4o and our model's counting performance by different object numbers.}
    \vspace{-10pt}
    \label{fig:GPT_bad_counting}
\end{figure}

\subsection{Intriguing findings}
Finally, we draw some intriguing ``byproduct'' findings from Table~\ref{tab:main_result_in_domain}, which can improve our understanding of the proprietary GPT  and Gemini models.\\
\textbf{GPT models struggle to count.} \taskCounting\ requires a model to count objects in an image (between 1 and 12), and GPT-4o can hardly count beyond 5 (see Figure~\ref{fig:GPT_bad_counting}).\\
\textbf{The \ourmodel\ can match or outperform GPT-4o on some cognitive tasks.} On \taskSpatialDetails\ and \taskWhoHasMore, the \ourmodel\ is on par with the four latest GPT and Gemini models. Moreover, it even outperforms GPT-4o on the math tasks of \taskCounting\ and \taskWhoHasMore. Figure~\ref{fig:GPT_bad_counting} shows that the \ourmodel\ counts better than GPT-4o given six or more objects.\\
\textbf{GPT vs.\ Gemini.} In general, the proprietary models give rise to similar results on \ourbenchmark. However, when we zoom into the individual tasks, GPT-5 is significantly better than the rest on \taskSpatialDetails, while Gemini models are better at \taskCounting\ than the GPT models. 

%% file: author-kit-CVPR2026-v1-latex-/tables/exp_in_domain.tex
\begin{table*}
\centering
\caption{\textbf{Performance comparison of different models on \ourbenchmark\ (in-domain).}
Different background colors denote different model families. 
We report accuracy (\%) for all tasks; the higher, the better.}
\vspace{-10pt}
\label{tab:main_result_in_domain}
\resizebox{\textwidth}{!}{
\begin{tabular}{lc!{\vrule}ccccccccccc}
\toprule
\multirow{2}{*}{\textbf{Model}} &
\multirow{2}{*}{\textbf{Overall}} &
\multirow{2}{*}{\textbf{Count}} &
\multirow{2}{*}{\textbf{LeftRight}} &
\multirow{2}{*}{\textbf{Spatial}} &
\multirow{2}{*}{\textbf{PV}} &
\multirow{2}{*}{\textbf{Memory}} &
\multirow{2}{*}{\textbf{Localization}} &
\multicolumn{3}{c}{\textbf{Visual Delay Response}} &
\multicolumn{2}{c}{\textbf{Who Has More}} \\
\cline{9-11} \cline{12-13}
 &  &  &  &  &  &  &  & \textbf{binary} & \textbf{multi-exact} & \textbf{multi-adjacent} & \textbf{synthetic} & \textbf{naturalistic} \\
\midrule

\rowcolor{groupC}
\multicolumn{2}{l!{\vrule}}{\textbf{Upper bound}} & \multicolumn{11}{l}{} \\
\rowcolor{groupC}
\texttt{Human performance}  &  {\tt 93.0} & {\tt 99.1}  & {\tt 94.5} & {\tt 100} & {\tt 91.8} & {\tt 97.9} & {\tt 87.3} & {\tt 98.2} & {\tt 63.6} & {\tt 95.5} & {\tt 98.2} & {\tt 96.4} \\[5pt]
\midrule
\rowcolor{groupA}
\multicolumn{2}{l!{\vrule}}{\textbf{Proprietary models}} & \multicolumn{11}{l}{} \\
\rowcolor{groupA}
Gemini-2.5-flash \cite{comanici_gemini_2025}  &  72.7  &  71.1  & 34.9 & 73.8 & 91.2 &  96.9 & 84.8 & 75.9  & 42.4 & 70.3 & 87.5 & 70.7 \\
\rowcolor{groupA}
GPT-4o \cite{openai_gpt-4o_2024} &  74.6 & 39.0 & 89.8 & 92.6 & 93.7 & 99.7  & 81.7 & 64.2 & 29.3 & 62.9 & 87.9 & 79.3 \\
\rowcolor{groupA}
Gemini-2.5-pro \cite{comanici_gemini_2025} &  82.5 & {\bf 77.2}  &  68.8 & 90.5 & 93.8 & 97.8 & {\bf 88.8} &  86.9  & 54.0 & 87.7 & {\bf 90.6} & 71.7 \\
\rowcolor{groupA}
GPT-5 \cite{singh2025openaigpt5card}  & {\bf 87.6}  &  69.1 & {\bf 96.0} & {\bf 94.5} & {\bf 95.0} &  {\bf 99.9} & 85.2 & {\bf 95.1} & {\bf 62.9} & {\bf 90.1} & 88.9 & {\bf 86.6} \\[5pt]
\midrule
\rowcolor{groupB}
\multicolumn{2}{l!{\vrule}}{\textbf{Open-source models - Bigger Size}} & \multicolumn{11}{l}{} \\

\rowcolor{groupB}
LLaVA-OneVision-7B \cite{li2024llavaonevisioneasyvisualtask} & 53.4  & 57.7 & 35.8 &  41.7 & 60.9  &  36.9  & 79.9 & 60.6 & 12.7  &  63.0 & 82.2 & 56.2 \\

\rowcolor{groupB}
Qwen2.5-VL-7B  \cite{bai_qwen25-vl_2025} &  63.7  &  44.1 &  32.5  & 79.5  &  84.5 &  35.2  & 85.6 & 73.8 & 39.6 & 69.0 &  92.1 & 64.7 \\

\rowcolor{groupB}
Qwen3-VL-30B-A3B \cite{yang2025qwen3technicalreport}  &  78.0  &  47.4 &  67.2  & 90.4  & 92.5  &  99.8  & 90.8 & 74.8 & 45.6 & 81.6 & 98.0  & 70.4 \\
\rowcolor{groupB}
\multicolumn{2}{l!{\vrule}}{\textbf{Open-source models - Similar Size as Ours}} & \multicolumn{11}{l}{} \\
\rowcolor{groupB}
LLaVA-OneVision-0.5B \cite{li2024llavaonevisioneasyvisualtask}   &  33.2 &  43.5 &  33.7  &  28.7 & 23.5  & 24.0  & 12.3 &  58.9 &  7.31 &  49.2 & 37.3  & 46.2 \\

\rowcolor{groupB}
InternVL3.5-1B \cite{Wang2025InternVL35AO}  &  37.2  &  27.9  &  32.2  &  34.6  &  34.4  &  25.8  &  44.8 &  64.1  &  11.6  &  36.8  &  47.8  &  49.1  \\

\rowcolor{groupB}
Qwen2.5-VL-3B \cite{bai_qwen25-vl_2025}   &  47.0 &  29.2  & 33.7  &  40.0 & \ul{71.7}  &  36.5 & \ul{85.8}  & \ul{66.7} &  17.0 & 32.7  &  51.7 & 52.3 \\

\midrule

\rowcolor{groupD}
\multicolumn{2}{l!{\vrule}}{\textbf{Ours}} & \multicolumn{11}{l}{} \\
\rowcolor{groupD}
\ourmodel    &  \ul{63.9} &  \ul{47.3} & \ul{96.4}  & \ul{92.8}  & 32.4 & \ul{90.8}  & 37.8  & 54.6  & \ul{38.1}  &  \ul{52.5} & \ul{99.7}  &  \ul{60.5} \\
[5pt]

\midrule
\rowcolor{groupE}
\multicolumn{2}{l!{\vrule}}{\textbf{Lower bound}} & \multicolumn{11}{l}{} \\
\rowcolor{groupE}
{\tt Random guess} & {\tt 31.8} & {\tt 8.33} & {\tt 33.3} & {\tt 33.3} & {\tt 25.0} & {\tt 25.0} & {\tt 25.0} & {\tt 50.0} & {\tt 12.5} & {\tt 37.5} & {\tt 50.0} & {\tt 50.0} \\

\bottomrule
\end{tabular}
}
\end{table*}

%% file: author-kit-CVPR2026-v1-latex-/tables/exp_synthetic_utterance.tex
\begin{table*}
\centering
\caption{\textbf{Two language sources for pretraining}.
\ourmodel-original is pretrained on our pretraining set whose language is mainly caregivers' speech transcripts, while \ourmodel-synthetic is pretrained on synthetic utterances generated by GPT-4o.}
\vspace{-10pt}
\label{tab:synthetic_utterance}
\resizebox{\textwidth}{!}{
\begin{tabular}{lc!{\vrule}ccccccccccc}
\toprule
\multirow{2}{*}{\textbf{Model}} &
\multirow{2}{*}{\textbf{Overall}} &
\multirow{2}{*}{\textbf{Count}} &
\multirow{2}{*}{\textbf{LeftRight}} &
\multirow{2}{*}{\textbf{Spatial}} &
\multirow{2}{*}{\textbf{PV}} &
\multirow{2}{*}{\textbf{Memory}} &
\multirow{2}{*}{\textbf{Localization}} &
\multicolumn{3}{c}{\textbf{Visual Delay Response}} &
\multicolumn{2}{c}{\textbf{Who Has More}} \\
\cline{9-11} \cline{12-13}
 &  &  &  &  &  &  &  & \textbf{binary} & \textbf{multi-exact} & \textbf{multi-adjacent} & \textbf{synthetic} & \textbf{naturalistic} \\
\midrule

\ourmodel-original  &  63.9 &  47.3 & 96.4  & 92.8  & 32.4 & 90.8  & 37.8  & \textbf{54.6}  & \textbf{38.1}  &  \textbf{52.5} & \textbf{99.7}  &  60.5 \\
\ourmodel-synthetic   &  \textbf{65.4} &  \textbf{52.6}  &  \textbf{99.3} &  \textbf{97.5} & \textbf{34.3} &  \textbf{99.9} & \textbf{38.3}  & 53.8  & 33.6 &  45.8  & 99.6  &  \textbf{64.7}  \\
\bottomrule
\end{tabular}
}

\end{table*}


%% file: sec/conclusion.tex
\section{Conclusion}
We introduced \ours, a framework that features a developmentally plausible pretraining set derived from the longitudinal SAYCam corpus, a compact VLM trained from scratch, and comprehensive developmental benchmarks (\ourbenchmark). \ourbenchmark\  adapts all vision-related measures from the newly published \nihtoolbox. It contains ten tasks spanning three subdomains (language, executive function/memory, and math) and requires a flexible model interface that can process image, video, and multi-turn dialogue. We demonstrate the potential of developmentally plausible vision FMs through extensive experiments on our pretraining and instruction tuning datasets, and we confirm the quality of \ourbenchmark\ through extensive benchmarking with proprietary and open-source models. This framework will serve as a principled platform to broaden research engagement in vision FMs and accelerate progress toward developmentally plausible learning. 

\medskip
\noindent\textbf{Acknowledgements.} We sincerely thank Jessica Sullivan, Chen Yu, and Michael C. Frank for their guidance on data use and insightful feedback on the project, which was partially supported by Sony Faculty Innovation Award, Gemini Academic Program Award, and Azure Sponsorship.

%% file: sec/X_suppl.tex
\clearpage
\def\thesection{\Alph{section}}
\def\theHsection{\Alph{section}}

\setcounter{section}{0}
\setcounter{table}{6}
\setcounter{figure}{6}

\maketitlesupplementary

\footnotetext[1]{The training dataset, the model checkpoints, the training scripts and the evaluation samples will be released to the public in the near future.}

\section{Model training}
\label{app:model-training}
\subsection{\ourmodel\ architecture}
We build upon the original BabyLLaVA-Llama model introduced in \babyvlm~\cite{wang_babyvlm_2025}, by giving it the capability to process multiple images as input and conduct multi-turn visual–linguistic interactions. The model architecture consists of a compact language backbone, a visual encoder, and a lightweight multilayer perceptron (MLP) connector that projects visual features into the language space. Unlike \babyvlm, which also experimented with smaller backbones (GPT-2~\cite{radford_language_nodate} + ResNeXt-50~\cite{xie2017aggregated}), we only adopt the larger variant composed of a LLaMA-1.1B~\cite{touvron_llama_2023, zhang_tinyllama_2024} language model and a ViT-L-16~\cite{dosovitskiy_image_2021} visual encoder (300M params). We find that the smaller variant often struggles to complete complex downstream tasks such as memory, primarily due to its limited model capacity, whereas our current configuration achieves a better balance between developmental plausibility and expressive capability. We also train a larger variant which doubles the size of the language model, and no significant gain is observed; thus, we only use this configuration for the paper.

\subsection{\ourmodel\ training paradigm}
\label{sec:training_paradigm}

We train the entire model from scratch using a four-stage pipeline, as summarized in Table~\ref{tab:training_paradigm}.

\input{author-kit-CVPR2026-v1-latex-/tables/training_paradigm}

\noindent\textbf{Stage 0: Unimodal Training.}  
In the first stage, the language and vision backbones are trained independently to acquire the basic representational abilities for each modality. The language backbone is trained on all transcribed utterances using a standard autoregressive loss~\cite{radford_improving_nodate}. Its tokenizer is initialized via Byte-Pair Encoding (BPE) \cite{bpe-sennrich-etal-2016} trained on the same corpus, with a fixed vocabulary size of 6000. The vision backbone is trained using a DINOv2 \cite{oquab_dinov2_2024} objective on SAYCam frames. We do not apply any filtering during this stage—except restricting samples to the training split—since the filtering procedures are primarily designed to enforce image–utterance alignment, which is irrelevant to unimodal representation learning.

\noindent\textbf{Stage 1: Feature Alignment.}  
This stage corresponds to Phase~1 training in LLaVA~\cite{liu_visual_2023}. Both the vision and language backbones are frozen, and only the MLP connector is optimized using an autoregressive loss. The objective is to align visual features with the language embedding space, effectively bridging the two modalities. To maintain training stability, we use only the image–utterance subset of the pretraining data in this stage, postponing exposure to multi-image inputs until later phases.

\noindent\textbf{Stage 2: Joint Pretraining.}  
In this stage, the vision backbone remains frozen, while the MLP connector and language backbone are trained jointly on the full mixed-format pretraining dataset, as described in Section~\ref{sec:training-set}. This allows the model to learn multimodal grounding over diverse input structures.

\noindent\textbf{Stage 3: Instruction Fine-tuning.}  
Finally, we fine-tune the model using the mixed instruction dataset, which is a combination of all the instruction samples mentioned in Table~\ref{tab:benchmark-mapping}. This step enables the model to perform various downstream tasks through natural-language prompts. The vision backbone, MLP connector and language backbone are all updated to learn instruction-following behavior and context-dependent reasoning. We apply two different learning rates for different modules in this stage: the learning rate of the vision backbone is 1e-4, while that of the MLP connector and language backbone is 5e-4.

Main hyperparameters of all 4 stages are summarized in Table~\ref{tab:training_paradigm}. All experiments are conducted on four NVIDIA A6000 GPUs with 48\,GB of VRAM each. Language backbone training completes in less than one hour, while the vision backbone completes in 4 days. Next, training the MLP connector requires approximately 5 hours. Joint pretraining on the mixed-format dataset takes roughly 34 hours to converge. Finally, instruction tuning takes $\sim$24 hours.

\subsection{Open-source model fine-tuning}

We conduct LoRA finetuning experiments on two open-source models, LLaVA-OneVision-7B and Qwen2.5-VL-7B, to evaluate the effectiveness of our instruction-finetuning dataset. Each task is finetuned separately. We set the LoRA rank to 64, use a scaling factor of 64, and apply a dropout rate of 0.05. Training is performed for 5 epochs with a global batch size of 128, a learning rate of 1e-4, a weight decay of 0.1, a warmup ratio of 0.03, and a cosine learning-rate schedule.

\section{Developmentally aligned benchmarks}
\label{appedix:benchmarks}
In Appendix~\ref{appedix:benchmarks}, we adopt the following organization: Subsection \ref{sec:common} describes general implementation details that are shared by several tasks, including details on the vocabulary used in \ourbenchmark, acquisition of SAYCam annotations, acquisition of Ego4d annotations, and important distinction between SAYCam and Ego4d. Then, each subsection between 2 and 11 describes how these annotations are used to construct one task in \ourbenchmark\ each, and are each broken up into \textit{Original Toolbox Task, Adaptation, Data Collection, and Example Prompt}. Some of these also include information on \textit{Evaluation} or \textit{Data Composition}.
\subsection{Data collection procedures common to all tasks} 
\label{sec:common}
\noindent\textbf{Vocabulary filtering}\\
\label{sec:vocab}
To ensure that all benchmarks in this work focus on developmentally appropriate vocabulary, we draw on the \textit{MacArthur–Bates Communicative Development Inventories (MAB–CDI): Words and Gestures} \cite{marchman_macarthur-bates_2023}. The MAB–CDI is a standardized instrument assessing early vocabulary comprehension and production in infants and toddlers, covering familiar words across core semantic categories (e.g., animals, foods, body parts, actions). 

Because it is widely regarded as a gold-standard reference for early lexical development, we restrict our benchmark vocabulary to words that appear in—or are closely aligned with—those in the MAB–CDI. Accordingly, during visual concept mining from SAYCam and Ego4D, we retain only crops whose labels fall within this developmentally grounded lexical domain, ensuring that every keyword used across tasks reflects concepts young children could plausibly understand.\\

\label{saycam-general}

\label{sec:annotations}

\noindent\textbf{SAYCam annotations}\\
To support all SAYCam-based benchmarks in this work, we build the following unified preprocessing pipeline that extracts high-quality image crops for every object and action concept appearing in the corpus. This pipeline is reused (with task-specific modifications described in the corresponding benchmark sections) across tasks and provides consistent visual grounding for all downstream datasets.\\
\begin{itemize}
    \item \textbf{Frame-level detection and indexing:}
    We first sample SAYCam videos at 1~FPS and run an open-vocabulary detector
(Grounding--DINO \cite{liu_grounding_2024}) using the GPT-annotated labels associated
with each frame as the open set. Let $\mathcal{S}$ denote the set of all such
SAYCam labels. For each label $s \in \mathcal{S}$, we construct an index
$\mathrm{Index}(s)$ that maps $s$ to all frames in which it is detected, together
with its proportionally buffered bounding boxes and GPT-derived blurriness scores.
This $\mathrm{Index}(s)$ structure serves as the master lookup table for retrieving
visual instances of any concept.

    \item \textbf{Normalizing label variants:} 
    Raw SAYCam labels $s \in \mathcal{S}$ often include plural forms, paraphrases, or
compositional descriptions. To ensure consistent visual grounding, we cluster
lexically or semantically equivalent labels into small groups based on lexical similarity, plural equivalence, and phrase containment heuristics. Each label $s$
is assigned to its cluster $\mathcal{M}(s)$. This allows us to treat
variants of $s$ such as “shoes”, “a shoe”, or “pair of shoes” as a single underlying
concept by retrieving visual instances from $\{\mathrm{Index}(s')| s' \in \mathcal{M}(s)\}$.

\item \textbf{Quality filtering:}
Because SAYCam contains naturalistic video frames from children's head-cam footage, many detections are of low-quality due to motion blur, wrong/irrelevant detector predictions, small/partial bounding boxes. Therefore, we score each detection result using four broad signals:
(1) detector confidence,
(2) CLIP image–text alignment,
(3) crop size, and
(4) spatial clarity (e.g., centeredness).
These signals are normalized per-concept and combined into a single quality measure. We also employ additional light-weight adjustments to ensure that within $\mathcal{M}(s)$, rare labels are not overwhelmed by frequent ones and that exact label matches are preferred over looser variants.

\item \textbf{Ensuring lexical and visual diversity:}
To avoid selecting many near-duplicate frames of the same scene, we apply simple diversity controls. We first ensure that different lexical variants of a concept are represented, and then enforce a minimal temporal spacing between chosen frames. From this diversified pool, we keep only a small number ($\leq10$) of final crops per concept, prioritizing clarity and representativeness.

\item \textbf{Final output:}
The result is a compact, high-quality set of image crops for every object or action concept in SAYCam. These curated crops act as the visual foundation for the majority of benchmarks built from SAYCam in this paper. They guarantee concept fidelity, diversity of visual contexts, and consistent quality standards across tasks.\\
\end{itemize}

\noindent\textbf{Ego4D Annotations}\\
For Ego4D, we do not perform any heavy data cleaning or processing due to the native, high-quality annotations of the dataset. 
For most of our benchmarks, we use image data from the \texttt{egotracks} split, which contains densely annotated egocentric video tracks. In addition, \taskPictureVocabulary\ (see Section~\ref{sec:pvts}) also draws image crops from \texttt{fho\_lta}—a subset of Ego4D focused on future hand–object interactions—providing additional object-centric visual diversity.

\noindent\textbf{Overall differences between SAYCam and Ego4d}\\
Here, we analyze the differences between SAYCam and Ego4d which result in \ourmodel's very poor generalization to Ego4d. Specifically, SAYCam was filmed by 3 babies across 4 homes, while Ego4d was filmed by 923 participants across 74 sites. There's also a significant domain shift in the size of the frames and the sizes of the objects relative to the frames.
See Table \ref{tab:comparison} for a summary.

\input{author-kit-CVPR2026-v1-latex-/tables/comparison}

In addition, although all of the Ego4d examples constructed in \ourbenchmark\ are directly based on objects listed in SAYCam's vocabulary, their backgrounds may still include objects that \ourmodel\ never saw in its training, and thus detract from its' overall understanding of the scene. For example, we might construct an example from Ego4d that asks about the location of a \textit{hand}, and although \ourmodel\ saw examples of \textit{hand} during training, the frame is full of other objects to which \ourmodel\ can attribute no meaning. In such a case, context clues learned by \ourmodel\ about where gloves are usually found relative to their scene, such as at the end of an arm or holding onto a known object, are lost, and performance drops correspondingly.
Further, we conjecture that this lack of generalization stems from not only the explicit action categories included in Ego4d that a baby would never have seen (like fixing a car or performing a laboratory experiment), but also from the inherently wider field of view captured adult demonstrators relative to babies. Further, we argue that even if Ego4d had been filmed of the same locations and actions as SAYCam, we would still observe a domain shift caused solely because the demonstrators are adults, perceiving the world from a higher point of view than babies. This point reinforces the uniqueness of the baby domain in the space of egocentric computer vision. \\


\subsection{Picture Vocabulary}\label{sec:pvts}
\noindent\textbf{Original Toolbox Task}\\
Our task is directly adapted from the \nihtoolbox~Picture Vocabulary Test (PVT), which evaluates a participant’s receptive vocabulary by presenting a spoken target word alongside four images (one correct, three distractors) \cite{gershon_nih_2024}. The goal is to touch the picture matching the target word. Distractors in the original PVT are designed to be \emph{plausible but incorrect}, typically encompassing coarse-categorical, fine-categorical, or phonological similarity. While the full PVT, taken directly from the NIH Toolbox\textsuperscript{\tiny\textregistered}~\cite{han_national_nodate, doi:10.1212/WNL.0b013e3182872e5f}, includes 373 examples, we identify 52 examples intended for early childhood receptive vocabulary evaluation through combining \texttt{all-MiniLM-L6-v2} embedding similarity \cite{10.5555/3495724.3496209} comparison to vocabulary in \cite{marchman_macarthur-bates_2023} and manual inspection.

While the Baby Toolbox PVT uses an IRT-based computer-adaptive score that converts response patterns into age-normed ability estimates \cite{han_national_nodate}, our adaptation simplifies this to straightforward 4-way accuracy since all items in the benchmark are evaluated rather than adaptively selected.

Our adaptation preserves the original developmental intent while replacing controlled illustrations with naturalistic egocentric visual inputs (SAYCam/Ego4D), providing a grounded benchmark for modeling baby-level vocabulary comprehension in realistic developmental environments.\\

\noindent\textbf{Adaptation}\\
To adapt the original PVT design to naturalistic corpora, we first map MAB-CDI words $r \in \mathcal{R}$ to corpus vocabularies $\mathcal{S}$: GPT-annotated labels for SAYCam and native objects/actions labels for Ego4D. This produces a set of visually grounded targets $\mathcal{G}_r \subset \mathcal{S}$ for each CDI anchor $r$, forming a one-to-many mapping $r \!\rightarrow\! \mathcal{G}_r$.

We then analyze the 52 baby-level NIH PVT items to quantify the original distractor structure. We define three categories: fine-categorical, coarse-categorical, and phonological. We manually annotate every NIH distractor to one or more of these types accordingly. Note that the original PVT includes unrelated distractors and we exclude those given the difficulty in controlling the quality of unrelated distractors in naturalistic imagery. We obtain the unnormalized distractor-type weights:
\[
w_{\mathrm{coarse}} = 0.5643,\quad
w_{\mathrm{fine}} = 0.1472,\quad
w_{\mathrm{phon}} = 0.0321.
\]
Using these proportions, we construct corpus-specific distractor pools from the entire corpus $
\mathcal{S}$ (because the model is only required to identify the correct target concept, not to correctly recognize or label the distractors):

\begin{itemize}
    \item \textbf{Fine-categorical:} We use similarity scoring based on CLIP text embeddings \cite{radford_learning_2021}. For SAYCam, candidates above a similarity threshold of $0.7$ is considered belonging to the same fine-grained category while for Ego4D we use a quantile band $[0.997, 0.99973]$ to also filter out overly similar and thus indistinguishable words.
    \item \textbf{Coarse-categorical:} We use Kmeans clustering based on CLIP text embeddings (SAYCam: $K=100$; Ego4D: $K=150$).
    \item \textbf{Phonological:} We use Soundex-based string similarity for both datasets.
\end{itemize}

For each CDI anchor $r$, we select a ground-truth label $g \in \mathcal{G}_r$ and sample three distinct distractors from these pools using the weights. For SAYCam examples, we perform a final round of manual screening to filter out the infeasible examples, while Ego4D examples are filtered with a hybrid procedure combining \texttt{Gemini2.5-flash} checks with a lightweight manual review.\\

\noindent\textbf{Data Collection}\\
To produce high-quality 4-way visual choices, we collect image crops corresponding to every target and distractor label.

For SAYCam, because \taskPictureVocabulary\ requires extremely precise, semantically clear images, we modify the fully automated pipeline in Section~\ref{sec:annotations} with the following changes:
\begin{enumerate}
    \item Candidates come \emph{directly} from $\mathrm{Index}(g)$ where $g \in \mathcal{G}_r$ given anchor $r$.
    \item Human annotators manually filter irrelevant, ambiguous, or blurry crops and refine bounding boxes, replacing automated quality scores.
\end{enumerate}
This process yields a compact, high-precision crop inventory used for all SAYCam examples.

For Ego4D, for the objects, we use the bounding boxes from the \texttt{visual\_crop} field of the \texttt{EgoTracks} benchmark, applying a deterministic buffer (1.2$\times$ + 8px margin) and requiring a post-buffer normalized area $>0.03$.
For actions, we use the \texttt{fho\_lta} benchmark which contains abundant action annotations. As there are no explicit bounding box annotations, we sample frames from the middle $25\%$ of each action frame interval and apply minimal center-biased cropping to maintain clarity.  
For each label, we keep 10 candidates while preserving diversity and visual fidelity, and we apply a \texttt{Gemini2.5-flash} pass to eliminate unusable crops.\\

\begin{figure}
    \includegraphics[width=1\linewidth]{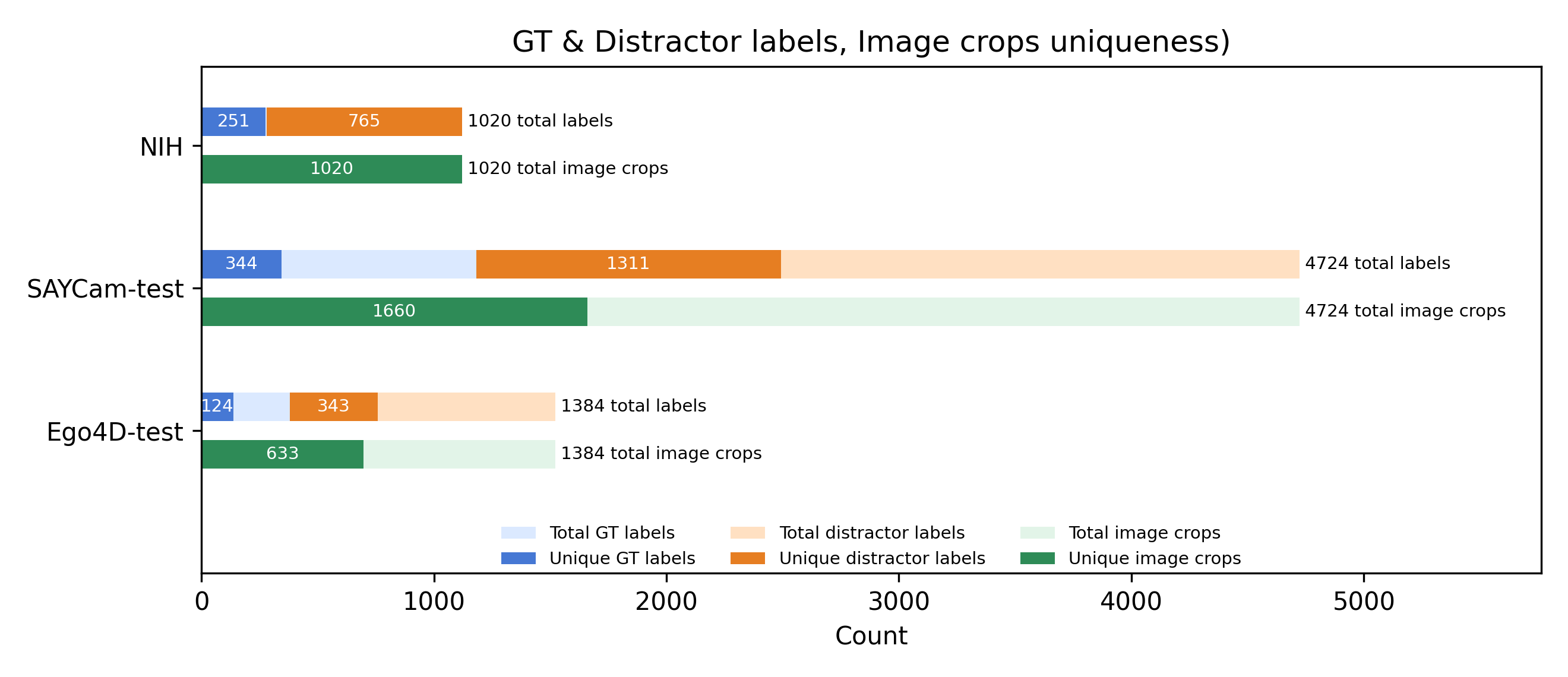}
    \caption{Label/Image crop uniqueness comparison between Picture Vocabulary Test in \nihtoolbox, SAYCam, and Ego4D.}
    \label{fig:unique}
\end{figure}
\begin{figure}
    \centering
    \includegraphics[width=1\linewidth]{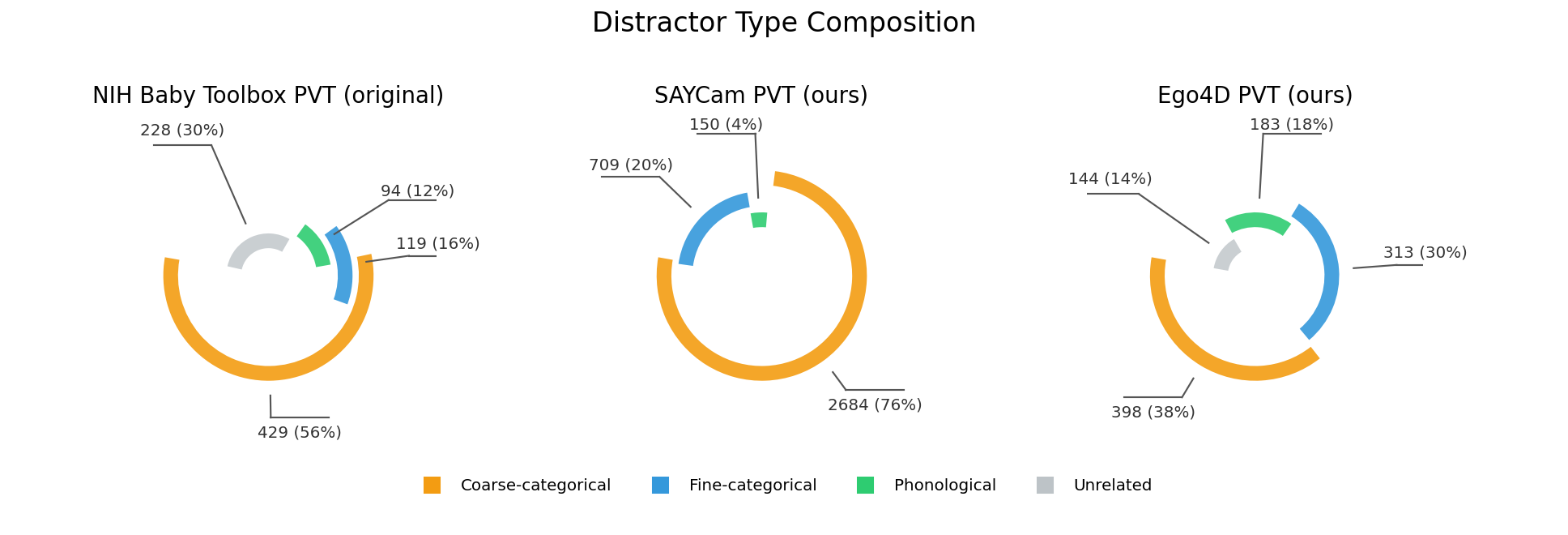}
    \caption{Distractor type composition for the Picture Vocabulary Test in \nihtoolbox, SAYCam, and Ego4D. The original PVT contains multi-type overlaps, while our sampling assigns each distractor a single type even though some satisfy multiple cues. Ego4D uses unrelated distractors only as a rare fallback.}
    \label{fig:pie}
\end{figure}


\noindent\textbf{Dataset composition}\\
As shown in Figure \ref{fig:unique}, We obtain 1181 SAYCam examples, covering 344 unique GT labels, 1311 unique distrator labels, and 1660 unique crops. Due to manual filtering, its distractor distribution only loosely follows NIH proportions (shown in Figure \ref{fig:pie}).
Similarly, we obtain 346 Ego4D examples over 124 unique GT labels, 343 unique distractor labels, and 633 unique images (shown in Figure \ref{fig:unique}) with the corresponding distractor distribution shown in Figure \ref{fig:pie}.\\

\noindent\textbf{Example Prompt}\\
Each finalized example is a prompt embedded with 4 image choices for which the following is an example:

\begin{quote}
\texttt{"
Touch the image of 'foot' (A) <image> (B) <image> (C) <image> (D) <image> "}
\end{quote}
The model needs to output one of \texttt{A}, \texttt{B}, \texttt{C}, or \texttt{D} to be evaluated.

\subsection{Looking While Listening}
\noindent\textbf{Original Toolbox Task}\\
The Looking while listening test (LwL) from \nihtoolbox\ aims to evaluate comprehension for object labeling and receptive language \cite{han_national_nodate}. The infant is shown two clipart images which is followed by an audio prompt describing one of them. Eye tracking is used to detect whether the participant is looking at the ground-truth image. Similar to PVT, we simplify the original metric to accuracy only.\\

\noindent\textbf{Adaptation}\\
To adapt LwL to our benchmark in SAYCam, We replace clipart with naturalistic image crops from SAYCam, and eye tracking with multiple choice, similar to \taskPictureVocabulary.\\  

\noindent\textbf{Data collection}\\
Examples for \taskLookWhileListen\ are taken directly from \taskPictureVocabulary\ examples.\\

\noindent\textbf{Example Prompt}\\
Each finalized example is a prompt embedded with 2 image choices for which the following is an example:

\begin{quote}
\texttt{"
Touch the image of 'foot'\\
(A) <image> (B) <image>"}
\end{quote}
The model needs to output one of \texttt{A} or \texttt{B} to be evaluated.\\

\subsection{Localization}
\noindent\textbf{Original Toolbox Task}\\
Much like Picture Vocabulary, the Mullen Receptive Language test \#19 tests infants on their ability to point at sketched \textit{target} objects as they are named, avoiding confusing them with the \textit{distractor} objects. Specifically, after gesturing to a group of sketched objects, the psychologist asks: \textit{Look at these. Where is the cat?} If the child points in the direction of the cat, they pass the test.\\

\noindent\textbf{Adaptation}\\
\taskLocalization\ makes a significant modification to the original \nihtoolbox\ measure- In \ourbenchmark, we find it meaningful to test pointing to objects \textit{in their naturalistic environments}, namely, we treat the objects naturally occurring in the background of the frame as distractors rather than inserting unrelated objects. Additionally, because if is infeasible to ask a model to 'point', the answer choices are always \textit{top left, top right, bottom left, bottom right}.

Again, the objects in this task are real objects from SAYCam and Ego4d rather than the sketches used in the \nihtoolbox, and just like in the \nihtoolbox, the prompt is the full frame and the name of the object to be localized.\\

\noindent\textbf{Data collection}\\
 The examples for both SAYCam and Ego4d are generated using the centers of the bounding boxes annotated in \ref{sec:annotations} and Ego4d's egotracks, respectively. 

\begin{figure}[t]
    \centering
    \includegraphics[width=1\linewidth]{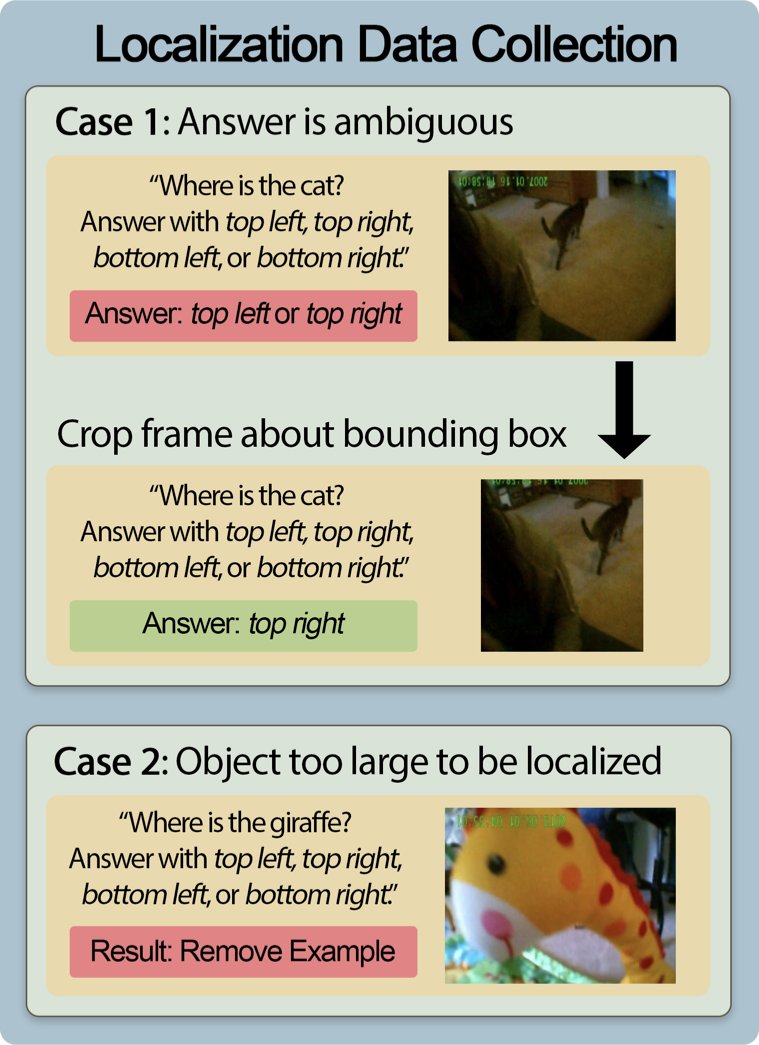}
    \caption{}
    \label{fig:localization_collection}
\end{figure}

To avoid including test examples where a bounding box stretches across two answers ambiguously (for example, an object in the bottom middle that could reasonably be called either \textit{bottom left} or\textit{ bottom right}), we 1) crop each frame so that its closest corner is flush with the edges of the object's bounding box, and 2) enforce a maximum bounding box area of 1/4 of the frame's area (see Figure \ref{fig:localization_collection}), which filters out 5.2k of the 7.3k possible test examples. In practice, we find that both of these steps are needed to ensure fair, reasonably unambiguous examples. We enforce no minimum confidence in the SAYCam object annotations and use all object names generated in \ref{sec:annotations}.\\

\noindent\textbf{Example Prompt}\\
Each finalized example is a prompt embedded with one image and the same four choices, for which the following is an example:

\begin{quote}
\texttt{"<image>\\
Point at the cup. Is it in (A) the top left of the image, (B) the top right, (C) the bottom left, or (D) the bottom right?"}
\end{quote}
The model needs to output one of \texttt{top left}, \texttt{top right}, \texttt{bottom left}, or \texttt{bottom right} to be evaluated.


\subsection{Left/Right}
\noindent\textbf{Original Toolbox Task}\\
\taskLeftRight\ is adapted directly from Mullen Visual Reception test \#29, in which a psychologist shows a child an object, then instructs the child to match it with the identical one. If the child correctly points to the identical object, avoiding confusing it with its own mirror image, they pass the test.\\

\noindent\textbf{Adaptation}\\
The only modification made while adapting VR test \#29 to \ourbenchmark\ is replacing the clipart objects with real objects from SAYCam and Ego4d. In \ourbenchmark, the basic format is preserved: a prompt image, followed by a correct answer and two distractor choices in some random order, are presented to the model. The target image is a duplicate of the prompt, and the incorrect answers are the mirror image of the target.

\begin{figure}[t]
    \centering
    \includegraphics[width=1\linewidth]{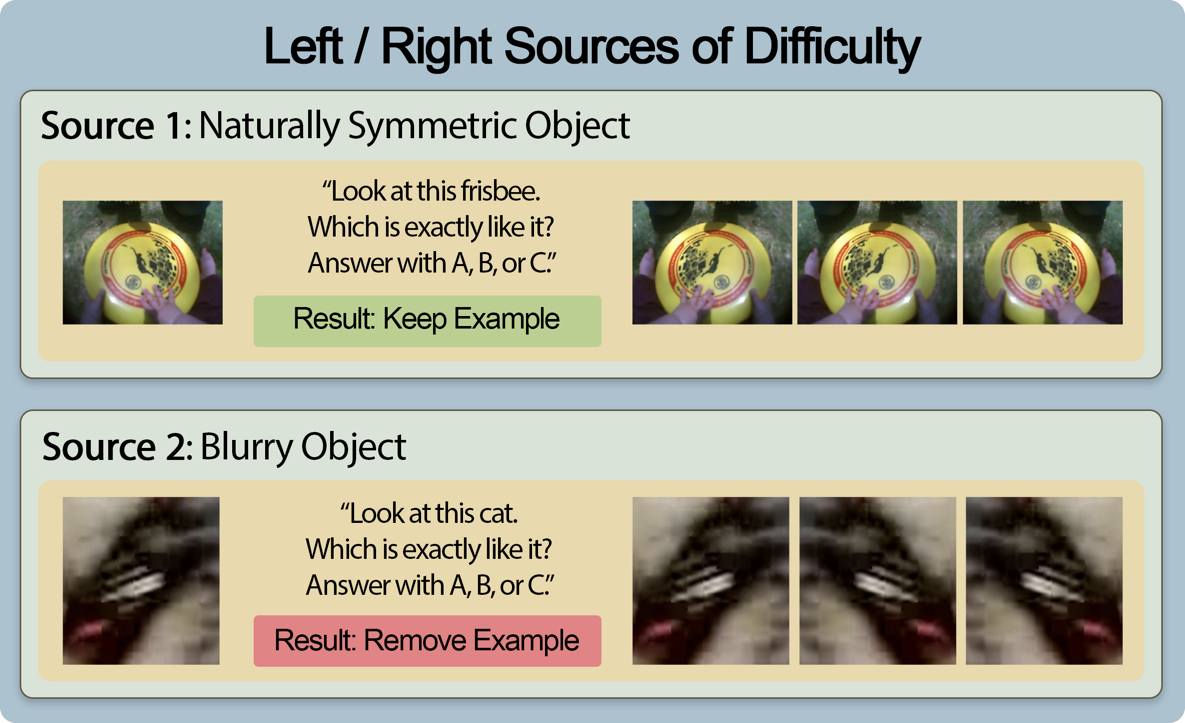}
    \caption{}
    \label{fig:LRdifficulty}
\end{figure}

Some examples in \taskLeftRight\ are harder than others; we conjecture that difficulty in this task can result from either 1) naturally symmetric objects, or 2) low resolution objects (see Figure \ref{fig:LRdifficulty}). Naturally symmetric objects are difficult because they require an encoding of fine-grained details. However, low resolution objects are difficult because even though there might be some spatial clues to discriminate the target from its mirror image, if have models can't ascribe any semantic meaning to the image, they won't encode any semantic meaning to its details. By filtering out small bounding boxes, we aim to remove the examples that are difficult solely due to low resolution. \\

\noindent\textbf{Data collection}\\
For the SAYCam variant, use the object names and bounding boxes generated in \ref{sec:annotations}. We enforce no minimum or maximum object size, and for the val and test splits, we enforce a minimum confidence in the bounding box of .85. In both variants, all object crops are zero-padded to (640, 480). 

For the Ego4d variant, we use object names and bounding boxes from the published Ego4d egotracks annotations and include only objects that belong to the vocabulary defined in \ref{sec:vocab}. To remove examples with poor resolution, we require either a minimum bounding box height or width of one fifth of the frame, which filters out about half of the otherwise qualifying examples.\\

\noindent\textbf{Example Prompt}\\
Each finalized example is a prompt embedded with 1 image prompt and 3 image choices for which the following is an example:
\begin{quote}
\texttt{"<image>\\
Which of the following is the same as this? (A) <image> (B) <image>, or (C) <image>?"}
\end{quote}
The model needs to output one of \texttt{A}, \texttt{B}, or \texttt{C} to be evaluated.

\subsection{Spatial Details}
\noindent\textbf{Original Toolbox Task}\\
Similarly, Mullen Visual Reception test \#25 also tests understanding of details in images. In this test, the child is presented with a sketch of a \textit{tulip}, and the psychologist asks: \textit{See this flower. Find one just like this. Look for it here}, while tracing their finger along a page filled with sketches of a \textit{tulip, a sunflower, a clover, and a daisy}. The child is allowed to refer back to the \textit{tulip} while choosing their answer. If the child points to the \textit{tulip}, they pass the test.\\

\noindent\textbf{Adaptation}\\
Again, the objects in our benchmark are real, cropped objects from SAYCam and Ego4d rather than clipart, and they come from more categories than just \textit{flower}. Additionally, because the models cannot "point" to the choices, the choices are passed as separate images and the correct answer is the index (A, B, or C) of the matching image. 

Our final modification to the original measure is that to make it more difficult for a computer, we present the answer choices in their naturalistic backgrounds rather than cropped as in the \nihtoolbox. In practice, we find the final modification necessary to make \taskSpatialDetails\ require a fine-grained understanding of detail, as matching identical images is trivial even for a small vision model. \\

\noindent\textbf{Data collection}\\
To construct examples from both SAYCam and Ego4d, we match objects with the label, but require that they come from different videos. The labels for each come from \ref{sec:annotations} and egotracks, respectively. Note that the same object can appear multiple times within an example- for instance, the same \textit{chair}, captured in two separate videos, can show up as two of the choices. In such cases, the model is forced to rely on spatial details such as orientation, perspective, and lighting, to match identical occurrences.

To ensure quality, we enforce a minimum object confidence of .92 in the SAYCam annotations. To increase difficulty, we also require that objects have an area of less than half of the frame's area.\\

\noindent\textbf{Example Prompt}\\
Each finalized example is a prompt embedded with 1 image prompt and 3 image choices for which the following is an example:
\begin{quote}
\texttt{"<image>\\
Which of the following is the same as this? (A) <image> (B) <image>, or (C) <image>?"}
\end{quote}
The model needs to output one of \texttt{A}, \texttt{B}, or \texttt{C} to be evaluated.

\subsection{Visual Delayed Response}

\noindent\textbf{Original Toolbox Task}\\
Inspired by Visual Delayed Response in the \nihtoolbox, we introduce an evaluation task designed to assess the spatiotemporal reasoning capabilities of vision-language models. More specifically, our task focuses on object tracking and spatial localization over time, requiring models to process multi-frame/video input to infer spatial trajectory and disappearance of a designated object. \\

\begin{figure}[t]
    \centering
    \includegraphics[width=1\linewidth]{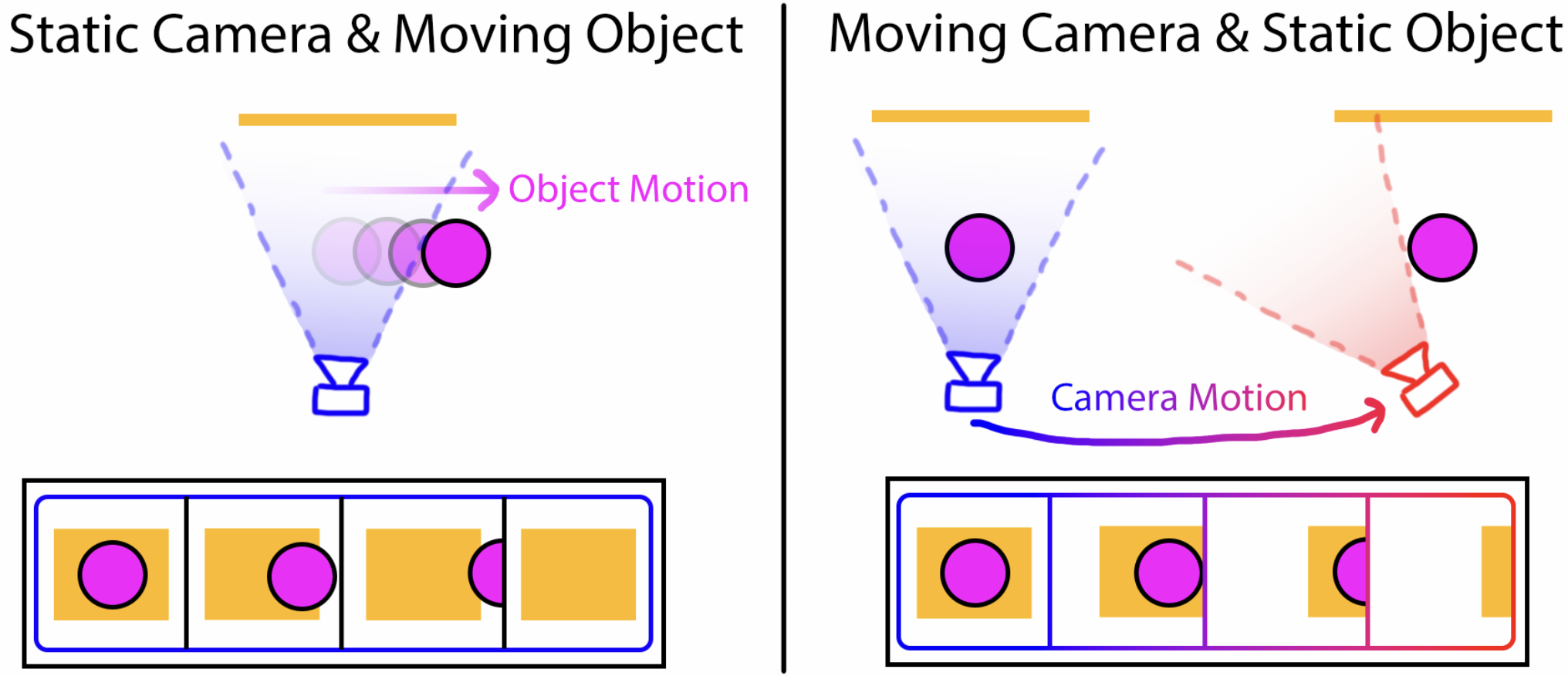}
    \caption{Comparison between sources of occlusion. \textbf{Left}: object occlusion from a static camera and moving object. \textbf{Right}: object occlusion from a moving camera and static object. Each panel shows a top-down view of the scene along with the corresponding projected 2D video depicting the occlusion event.}
    \label{fig:vdr_object_motion_vs_camera_motion2}
\end{figure}

\begin{figure}[t]
    \centering    \includegraphics[width=1\linewidth]{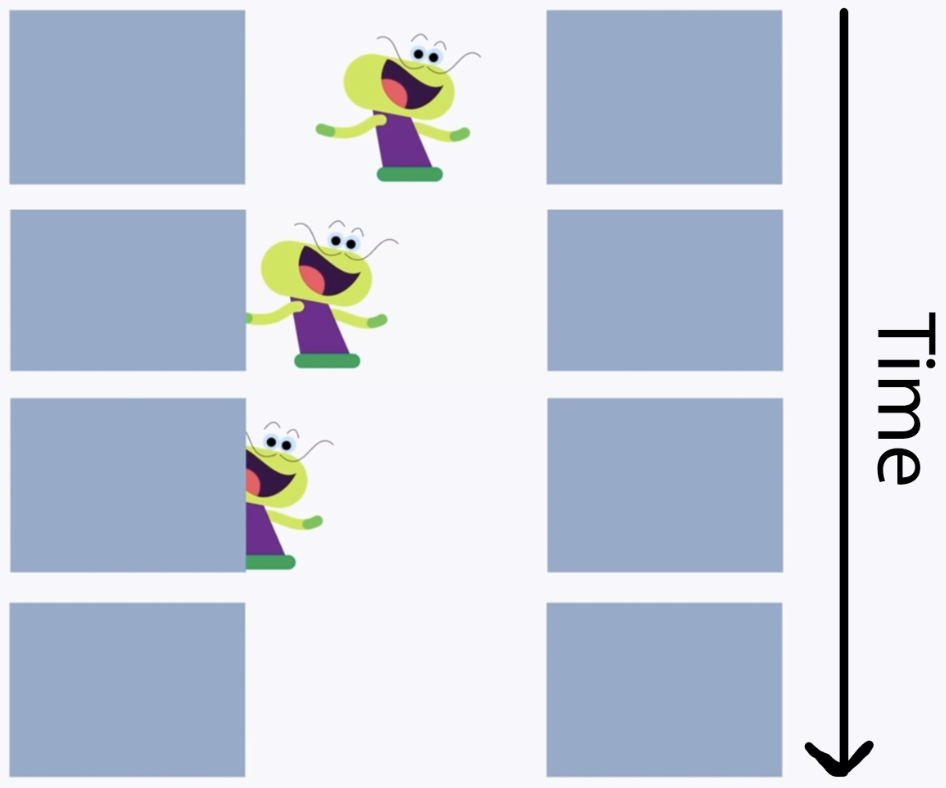}
    \caption{Example of Visual Delayed Response task, taken directly from the NIH Baby Toolbox.}
    \label{fig:vdr_nih_baby_toolbox_example}
\end{figure}

\noindent\textbf{Adaptation}\\
In the original \nihtoolbox\ task, a cartoon creature is placed in a frame with grey walls to its left and right. The creature moves from the center of the frame to behind either wall, and the child must identify which wall the creature hid behind. (See Figure \ref{fig:vdr_nih_baby_toolbox_example})

Translating this task to real-world videos is challenging, as the synthetic examples from the toolbox portray an unrealistically ideal scenario. Each toolbox example depicts a moving object observed from a static camera perspective, with simplified backgrounds and perfectly smooth motion trajectories. Such controlled scenarios are rare in real-world footage, especially in egocentric videos captured from a toddler’s perspective.

To address this challenge we exploit the frequent head movements captured in SAYCam footage, together with the fact that many objects in real-world scenes are largely stationary. By inverting the source of 2D object motion from a static camera with moving objects to a moving camera with stationary objects (see Figure \ref{fig:vdr_object_motion_vs_camera_motion2}), we are able to expand the dataset by over an order of magnitude.

Formally, the model is provided a video $V = \{f_1, f_2, ... f_T\}$ and designated key object $k$. The video depicts the key object $k$ moving within the field of view and eventually exiting the visible frame at time $t^* \leq T$. The model’s objective is to predict the exit region $r \in \mathcal{R}$, where $\mathcal{R}$ denotes the set of possible frame boundaries through which the object may leave. 

We define two variants of this task, which differ in the set of selectable exit regions $\mathcal{R}$ provided to the model:

\begin{itemize}[leftmargin=1em]
\item \textbf{Multi-choice setting}: $\mathcal{R}_m$ = $\{$left, right, top, bottom, top-left, top-right, bottom-left, bottom-right$\}$

\item \textbf{Binary setting}: $\mathcal{R}_b$ = $\{$correct, opposite$\}$, $\mathcal{R}_b \subseteq \mathcal{R}_m$ 
\end{itemize}

The multi-choice variant provides a comprehensive set of possible exit regions, where the model is given eight regions as selectable options. The binary variant is a simplified version of the task, where the model only chooses between two options: the correct exit region or the region directly opposite to it.

The overall task can be summarized as a mapping \(f_{VDR}(V, k) \rightarrow r\), where \( r \in \mathcal{R}\). Here, $f_{VDR}$ represents the function that, given a video $V$ and designated key object $k$, predicts the exit region $r \in \mathcal{R}$ through which the object leaves the frame. \\

\noindent\textbf{Data Collection}

\textbf{SAYCam.} Collecting examples for the SAYCam variant of \taskVDR\ can be split into 3 stages: filtering with GPT annotations, filtering with object tracking, and manual labeling. 

\text{Stage 1.} We first use the 1 FPS annotations provided by GPT in \ref{sec:annotations}, where each frame is labeled with a "key object" and "objects" attribute. The "key object" denotes a singular object being attended to in a particular frame (if any), and "object" denotes a list of all visible objects within the frame of view. We do an initial filtering for candidate clips by using a sliding window over the 1 FPS frames of each long-range video. For a clip to pass the filter, the first half of frames in the window must have the same "key object", $k$. In addition, the second half of frames must not have $k$ listed as a "key object" or be present in the "objects" list. From the 422990 initial clips, 17443 are passed as candidate clips to the next stage.

\text{Stage 2.} We then perform open-set object detection \cite{wang2025yoloerealtimeseeing} over the 1 FPS frames sampled from each candidate clip, where the only object class to be detected is the "key object" itself. An object tracking algorithm \cite{zhang2022bytetrack} is also used to track the "key object" over the full fps video. The clips are filtered according to the object tracks, where each track must satisfy all of the following:
\begin{itemize}[label=--, left=0em, labelsep=1em, align=parleft]
    \item Start within the middle 70\% of the frame
    \item Appear in at least 10 consecutive frames
    \item Disappear for at least 10 consecutive frames before the full clip ends
\end{itemize}
To help account for errors in the object detection/tracking, we purposefully loosen the filters and add additional measures for sporadic/false detections. From the 17443 initial clips, 3908 are passed as candidate clips to the next stage.

\text{Stage 3.} The final stage involves manually reviewing and hand-labeling each candidate example from the previous stage. We label not only for the ground truth exit direction, but also for a variety of annotations related to overall quality of the clip. In total, we annotate for camera motion, scene visibility, camera stability, occlusion, exit direction, and presence of multiple objects. A breakdown for each is provided as follows:

\begin{itemize}[label=--, left=0em, labelsep=1em, align=parleft]
    \item \textbf{Occlusion}: \{Fully Occluded, Partially Occluded, Remains in View\}
    \item \textbf{Camera Motion}: \{Static, Moving\}
    \item \textbf{Direction of Exit}: [Up, Down, Left, Right]
    \item \textbf{Scene Visibility}: \{Excellent, Good, Fair, Poor\}
    \item \textbf{Camera Stability}: \{Very Stable, Stable, Shaky, Very Shaky\}
    \item \textbf{Multiple Objects}: \{True, False\}
\end{itemize}

We then filter for valid high-quality clips according to the following criteria:

\begin{itemize}[label=--, left=0em, labelsep=1em, align=parleft]
    \item Object must become fully occluded
    \item Direction of exit cannot be contradicting (both left \& right, or both up \& down)
    \item Scene visibility better than "Poor"
    \item Camera stability better than "Very Shaky"
\end{itemize}

From the 3908 initial clips, 2380 are passed as final clips for the dataset.

\textbf{Ego4D.} Data collection for Ego4D follows a very similar structure to the SAYCam process, with the addition of tracked object annotations being already provided by the Ego4D dataset. We use a sliding window over each long-range video's object tracks and filter for all of the following:

\begin{itemize}[label=--, left=0em, labelsep=1em, align=parleft]
    \item Object is present in first half of window and disappears in second half
    \item Object bounding box $\geq$ 40000 pixels ($\sim$13\% of screen)
    \item Starts within the middle 50\% frame
\end{itemize}

Each clip is also manually reviewed/labelled according to the same procedure as SAYCam data collection

\textbf{Multi-frame versions.} Since the average clip can range from 100-150 frames, we manually create multi-frame counterparts to each example. More specifically, we look to obtain 1 representative object frame and 3-9 linearly sampled frames that best showcase the object motion/disappearance in a given clip. To do this, we first find the specific frame for three different fields: full object frame, start occlusion frame, and end occlusion frame. The full object frame is always used as the first frame in the multi-frame sequence, and shows the key object in clear view. The start/end occlusion frames mark the interval with which the key object becomes occluded. A random number of frames (3-9) are linearly sampled along this interval to complete the multi-frame sequence for a given clip. \\

\begin{figure}[t]
    \centering
    \includegraphics[width=1\linewidth]{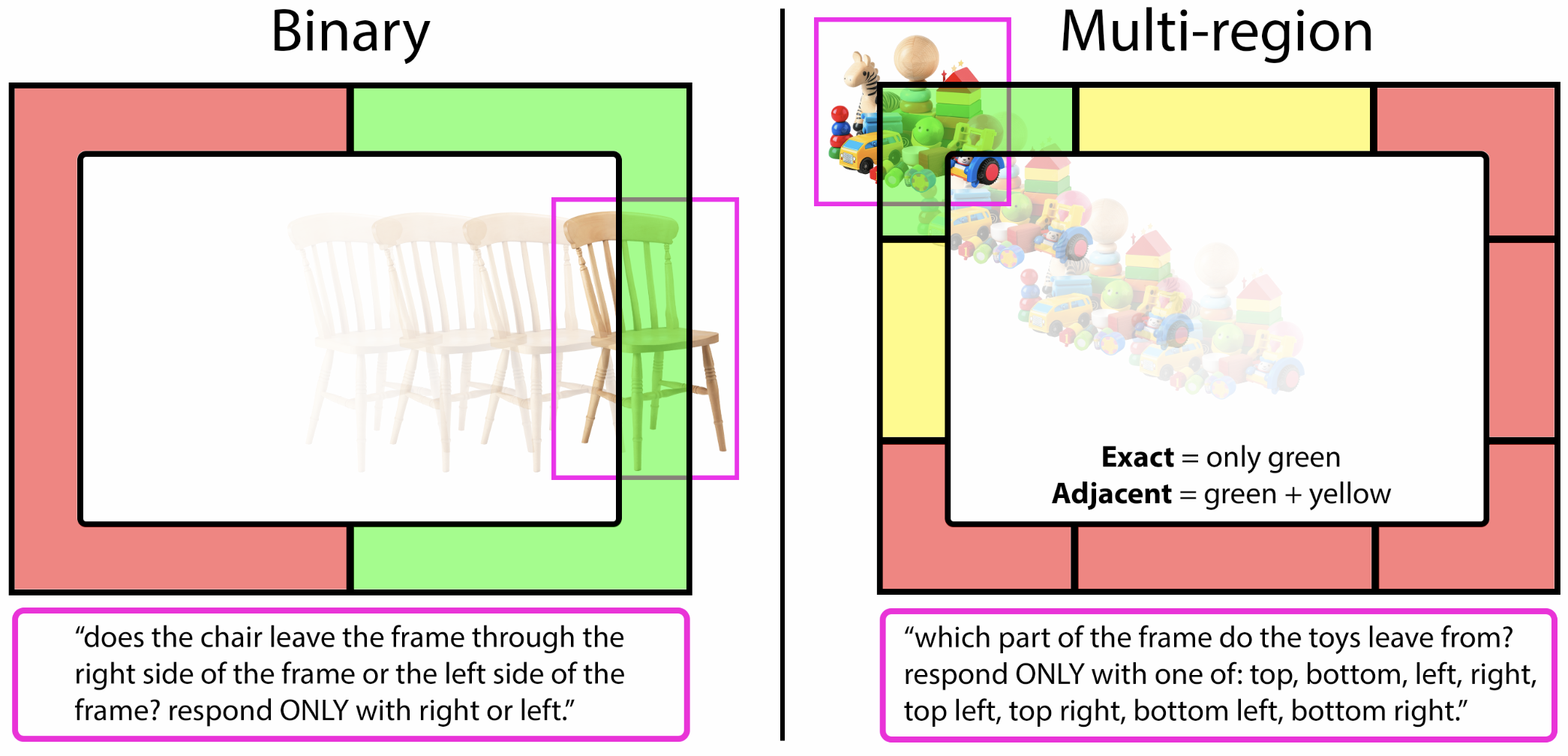}
    \caption{Visualization of evaluation methods for Visual Delayed Response task. \textbf{Left}: Binary evaluation for the binary setting, where there is only a correct and opposite incorrect option. \textbf{Right}: Exact and Adjacent evaluation for the multi-region setting, where the correct region for Exact is defined by only the green region, and the correct region for Adjacent is defined by both the green and yellow regions.}
    \label{fig:vdr_evaluation_methods2}
\end{figure} 


\noindent\textbf{Evaluation}\\
Evaluation is performed over three separate variants: \textbf{Exact} and \textbf{Adjacent} in the multi-choice setting, and \textbf{Binary} in the binary setting (see Figure \ref{fig:vdr_evaluation_methods2}). Accuracy is used as the metric for evaluation, defined as the fraction of predictions considered correct across all trials for a given variant. 

\noindent In the multi-choice setting:

\begin{itemize}
    \item \textbf{Exact}: Only the labelled ground truth region is counted as correct.

    \item \textbf{Adjacent}: Both the labelled ground truth region and its two adjacent regions are counted as correct. This helps account for small ambiguities in the ground truth label.
\end{itemize}

\noindent In the binary setting:

\begin{itemize}
    \item \textbf{Binary}: A prediction is correct if it matches the "correct" region rather than the "opposite" region. \\
\end{itemize}

\noindent\textbf{Example Prompt}\\
Each finalized example includes a series of \texttt{<image>} tags or singular \texttt{<video>} tag, followed by the prompt. To be properly evaluated, the model must output exactly one option from the choices given in the prompt.

\noindent Example from binary setting with multi-frame input:
\begin{quote}
\texttt{"<image><image><image><image> \\
does the bottle leave the frame through the right side of the frame or the left side of the frame? respond ONLY with 'right' or 'left'."}
\end{quote}
\noindent Example from multi-choice setting with video input:
\begin{quote}
\texttt{"<video>\\ which part of the frame do the toys leave from? respond ONLY with one of: 'top', 'bottom', 'left', 'right', 'top right', 'top left', 'bottom right', or 'bottom left'."}\\
\end{quote}

\subsection{Memory}
\begin{figure*}[t]
    \centering
    \includegraphics[width=\textwidth]{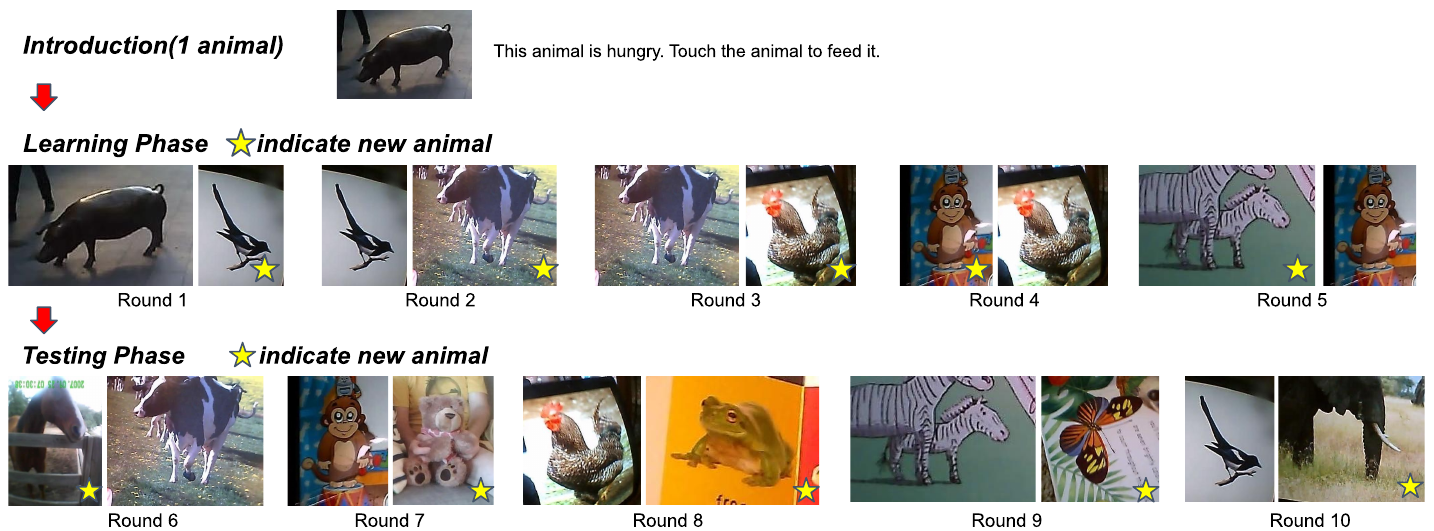}
    \caption{A sample of our memory task adaptation.
We use the MAB-CDI words detected in SAYCam as the images to be memorized.}
    \label{fig:memory-demo}
\end{figure*}

\noindent\textbf{Original Toolbox Task}\\
The Memory task in the NIH Toolbox is designed to measure how well toddlers (22–42 months old) learn and remember new information using a touchscreen. Children play a short game where they “feed” hungry cartoon animals by touching them on the screen. The test is divided into the learning phase and the test phase.
\begin{itemize}[leftmargin=1em]
\item \textbf{Learning phase:} children see pairs of animals and are told to touch the new animal—the one they have not fed before. They complete 10 trials and receive feedback so they can learn the rules and memorize the animals seen in this phase.

\item \textbf{Testing phase:} children again see pairs of animals and told to touch the new animal, where each old animal from the learning phase appears twice, each time paired with a different new animal. They complete 20 trials and receive no feedback so correct responses reflect their memory for animals in learning phase. 

\end{itemize}
 The animals were selected based on how many 24-month old infants were familiar with them according to data from the MB-CDI Wordbank. Performance is scored based on whether the child touches the correct animal in the testing phase, along with optional reaction time measures to show how quickly they respond.\\
 
\noindent\textbf{Adaptation}\\
\noindent
To simplify the problem and enlarge the potential dataset size, we define the set of word labels used in the learning phase as
\[
\mathcal{W}_{\text{learn}} = \{w_1, w_2, \dots, w_k\},
\]
where each \( w_i \) corresponds to an image \( x_i \in \mathcal{X}_{\text{learn}} \).
These image–label pairs \( (x_i, w_i) \) serve as the stimuli to be memorized during the learning phase.
We further sample \( 2k \) additional word labels for the testing phase,
\[
\mathcal{W}_{\text{test}} = \{w_{k+1}, w_{k+2}, \dots, w_{3k}\},
\]
each associated with a novel image \( x_j \in \mathcal{X}_{\text{test}} \).

\noindent
At each round \(t\), the Vision–Language Model (VLM) receives an input consisting of two images and a text prompt:
\[
I_t = \{x_{p_t}, x_{q_t}, P_t\},
\]
where \(x_{p_t}, x_{q_t}\) are the image inputs and \(P_t\) is the corresponding prompt.
\begin{itemize}

\item \textbf{Learning phase:}
The learning phase contains \(k\) rounds:
\[
I_t^{\text{learn}} =
\begin{cases}
\{x_1, P_1\}, & t = 1, \\[4pt]
\{x_{t-1}, x_t, P_t\}, & 2 \le t \le k,
\end{cases}
\]
where the two images in the second case are presented in random order.
This setup enables the model to incrementally associate visual concepts across consecutive rounds within a single context window.

\item \textbf{Testing phase:}
The testing phase consists of \(2k\) rounds, each comparing a learned stimulus with a new one:
\[
I_t^{\text{test}} = \{x_{i(t)}, x_{j(t)}, P_t^{\text{test}}\},
\quad
x_{i(t)} \in \mathcal{X}_{\text{learn}}, \;
x_{j(t)} \in \mathcal{X}_{\text{test}}.
\]
Here, \(x_{i(t)}\) is a previously seen image and \(x_{j(t)}\) a novel one. The model must identify which image corresponds to the new concept described in \(P_t^{\text{test}}\). \\
\end{itemize}

\noindent \textbf{Evaluation}\\
Each learned concept \( w_i \in \mathcal{W}_{\text{learn}} \) is paired with two distinct new concepts:
\begin{align}
(w_i, w_{a(i)}),\ (w_i, w_{b(i)}), & \quad
a(i), b(i) \in \{k{+}1, \dots, 3k\}, \nonumber\\
& a(i) \neq b(i),
\end{align}
forming two dyads per old stimulus and a total of \(2k\) dyads in the testing phase.
To mitigate the influence of random guessing, an old stimulus \( w_i \) is considered successfully \textit{remembered} only if both of its dyads are answered correctly:
\[
r_i =
\begin{cases}
1, & \text{if both dyads for } w_i \text{ are correct},\\[4pt]
0, & \text{otherwise.}
\end{cases}
\]
The overall memory accuracy is then computed as
\[
\text{Acc}_{\text{mem}} = \frac{1}{k} \sum_{i=1}^{k} r_i.
\]

\noindent
In all experiments, we set \( k = 5 \), resulting in a total of \(3k = 15\) distinct word–image pairs.
This design preserves the spirit of the original Toolbox while adapting the procedure to the VLM’s limited context window.
When designing the evaluation metric, we follow the structure of the original Toolbox with appropriate simplifications.
Specifically, we remove the original intermediate 6–8 min delay settings between the learning and testing phases in our benchmark design.
Future extensions may incorporate external memory mechanisms such as Retrieval-Augmented Generation (RAG),
or introduce irrelevant contexts between the two phases to simulate real-world temporal gaps.
In this work, however, we focus exclusively on assessing the model’s in-context retrieval ability. \\

\noindent\textbf{Data collection}\\
For the scalability of the memory task, we expanded the image set from the cartoon animals in the original Toolbox to the objects in the SAYCam dataset, which also ensures that the items are familiar to children. We used a combination of annotation-based search scripts and automated vision models, including CLIP for object–text similarity and SAM for object segmentation as shown in \ref{sec:annotations}, to find and isolate frames where these objects appeared clearly. Manual screening was also done after auto-filtering. This process allowed us to gather real-world visual examples of common objects seen by young children, supporting the creation of new learning and memory trials for our benchmark. The visual objects collected from SAYCam dataset will serve as our stimuli in the memory task. \\

\noindent\textbf{Example Prompt}\\
Each finalized example is a list of prompts each embedded with 2 image choices, for which the following is an example:

\begin{quote}
\texttt{"Let's try more.\\
Touch the new image.\\
 (A) <image> or (B) <image>."}
\end{quote}
The model needs to output one of \texttt{A} or \texttt{B} to be evaluated.\\

\subsection{Who Has More}
\noindent\textbf{Original Toolbox Task}\\
In the \nihtoolbox, the Who Has More Measure is poised as a simple narrative: there are two animals; each of them is pictured with some number of the same object. Which animal has more?\\

\noindent\textbf{Adaptation}\\
In \ourbenchmark, we remove the narrative aspect and replace the clipart objects with naturalistic SAYCam and Ego4d objects. In the \textit{Naturalistic} adaptation, the objects are not necessarily identical and appear in their naturalistic backgrounds; in the \textit{Synthetic} adaptation, the objects are perfectly identical, cropped, and pasted onto black backgrounds in matching layouts. The model is prompted to identify whether the \textit{first} or \textit{second} has more. \\

\noindent\textbf{Data collection}\\
In the synthetic variants, to pick the two quantities to compare, we first sample a number between one and ten. Then, from the numbers remaining that are \textit{lower than the first one}, we sample the second quantity. We do this to ensure a balanced distribution in the \textit{differences in numbers being compared} for each answer. The objects being compared come from the annotations in \ref{sec:annotations} and egotracks for SAYCam and Ego4d, respectively.

For the test sets in the naturalistic adaptations, each example is hand-annotated by two separate human experts to cross-validate annotation quality. Specifically, the first human expert labels video frames with an object type and the number of that object. Next, for each of the frames that the first annotator labeled, the second annotator labels the number of the named object in each, \textit{without access to the first annotator's annotation}. 

With both labels for each frame, we construct an example for every pair of frames of with objects of the same type for which \textit{both annotators would have arrived at the same answer answer as to which has more had they based their decision solely on their count annotation}. As an example, say the first annotator labels frame A as having 5 cups, and frame B as having 6 cups. If the second annotator labels 5 cups in frame A and 7 cups in frame B, we construct a \taskWhoHasMore\ example from frames A and B (despite the annotators giving frame B two different labels) because 5<7  \textit{and} 6<7. However, if the second annotator instead labeled frame B as having 5 cups, we \textit{do not} construct a \taskWhoHasMore\ example from frames A and B, because the two annotators would have given different answers for such an example.

In constructing \taskWhoHasMore, we observe that some objects occur in multiples more than others, and each object follows a unique (and usually nonuniform) distribution of quantity- for example, the number of hands visible in a frame is usually one or two and rarely another number, while an object like books could reasonably be seen in any quantity between one and ten. Additionally, we observe that given the differences in settings and scene perspective, the distributions of object types as well as quantity per object is inherently different for SAYCam and Ego4d.  \\

\noindent\textbf{Example Prompt}\\
Each finalized example is a prompt embedded with 2 image choices for which the following is an example:
\begin{quote}
\texttt{Which of the following has more of shoe? (A) <image>, or (B) <image>?"}
\end{quote}
The model needs to output one of \texttt{A} or \texttt{B} to be evaluated.\\

\subsection{Subitizing}
\noindent\textbf{Original Toolbox Task}\\
In the \nihtoolbox, the infant sees one to four colored dots for only one second, then an audio prompt requests the number of dots. Importantly, the dots are not shown for long enough to be counted one at a time- Subitize is intended to measure the ability to \textit{quickly identify small quantities, without counting.}\\

\noindent\textbf{Adaptation}\\
To construct \taskSubitizing\ in \ourbenchmark, we paste objects onto random locations on black frames, in random quantities between one and four. To simulate the "one second flash", we insert empty frames before and after the frame including the objects. \\

\noindent\textbf{Data collection}\\
In the SAYCam variant, the objects being pasted come from frames cropped by the bounding boxes obtained in Section \ref{sec:annotations}, subjected to a minimum confidence of .95. In the Ego4d variant, the bounding boxes come from egotracks, and only objects in the MAB-CDI vocabulary are included. \\

\noindent\textbf{Example Prompt}\\
Each finalized example is a prompt embedded with 1 blank frame, 1 image prompt, and 1 blank frame for which the following is an example:
\begin{quote}
\texttt{<image> <image> <image>\\ How many of apple did you see? Answer with 1, 2, 3, or 4."}
\end{quote}
The model needs to output one of \texttt{1}, \texttt{2}, \texttt{3}, or \texttt{4} to be evaluated.\\


\subsection{Object Counting}
\noindent\textbf{Original Toolbox Task}\\
In the \nihtoolbox, infants are shown some number of an object on a screen, and asked to count them. Unlike the Subitize measure, there is no time limit- participants have time to count each item individually.\\

\noindent\textbf{Adaptation}\\
In \ourbenchmark, the examples are constructed in the same way as the \taskSubitizing examples, except the quantities are between one and twelve, and there are no blank frames corresponding with the lack of a time limit.\\

\noindent\textbf{Data collection}\\
The data collection for \taskCounting is the same as for \taskSubitizing. \\

\noindent\textbf{Example Prompt}\\
Each finalized example is a prompt embedded with 1 image prompt, for which the following is an example:
\begin{quote}
\texttt{<image>\\ How many of chair did you see? Answer with a number 1-12."}
\end{quote}
The model needs to output a number between \texttt{1} and \texttt{12} to be evaluated.\\

\section{Human survey}
\label{app:humansurvey}


\subsection{Small-scale human adult test}

To confirm the validity of \ourbenchmark, we collect small-scale adult performance data on eight of the ten tasks. We omit \taskLookWhileListen\ and \taskSubitizing\ as their examples are directly taken from \taskPictureVocabulary\ and \taskCounting, respectively. In total, we have data from n=11 adult participants, each completing 10 trials per task for the SAYCam variants of \taskPictureVocabulary, \taskLocalization, \taskLeftRight, \taskSpatialDetails, \taskVDR, and \taskCounting, and 5 trials per task for the SAYCam variants of naturalistic \taskWhoHasMore\ and synthetic \taskWhoHasMore, and as well as the Ego4d variants of all tasks other than \taskMemory. Participants completed 30 consecutive rounds of each \taskMemory\ variant, requiring a maximum memory of 29 distinct images. 

Results for each task can be found in the \textit{Human performance} rows of Tables~\ref{tab:main_result_in_domain} and \ref{tab:main_result_ood}. In summary, our participants achieved an average accuracy of 93.0 on all SAYCam tasks and 93.5 on all Ego4d tasks, for both of which they far outperform any model. From this, we conclude 1) \ourbenchmark\ is a valid discriminator of vision FMs with adult performance as a strong upper bound, and 2) the SAYCam and Ego4d variants have roughly similar complexity and ambiguity for humans.\\

\subsection{Children Helping Science tests}

\begin{figure*}
    \centering
    \includegraphics[width=\linewidth]{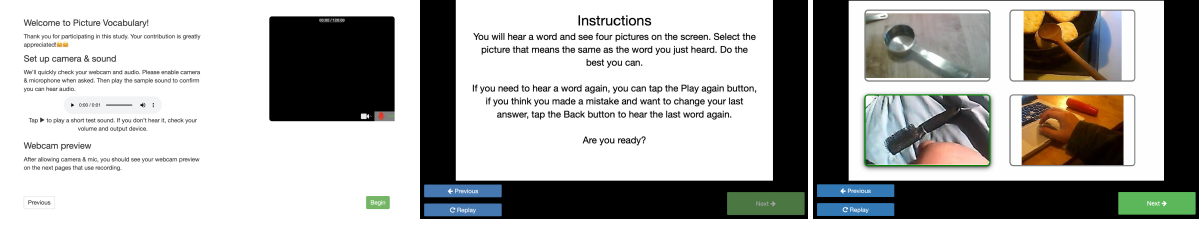}
    \caption{User interface design for our CHS-adapted Picture Vocabulary task.}
    \label{fig:pvt-ui}
\end{figure*}

\begin{figure*}
    \centering
    \includegraphics[width=1\linewidth]{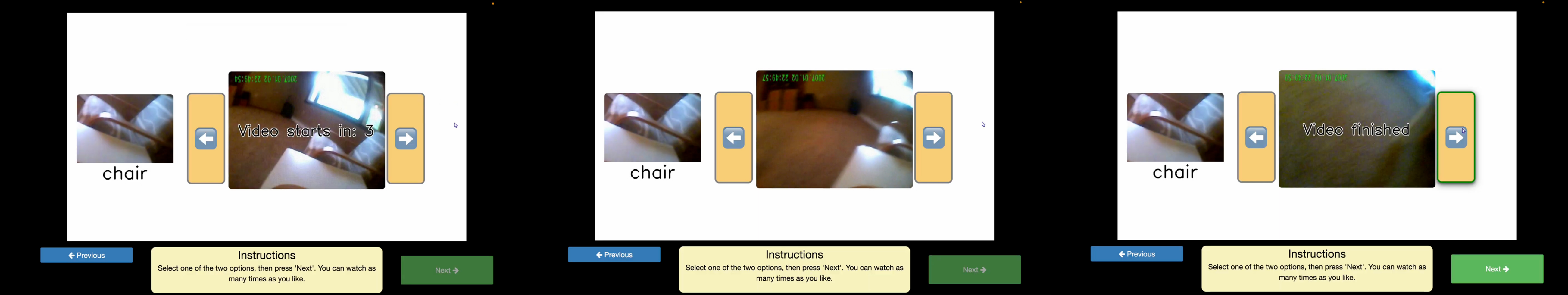}
    \caption{User Interface design for the trial page of Visual Delayed Response task on CHS.}
    \label{fig:vdr-ui}
\end{figure*}

To further examine the developmental fidelity of \ourbenchmark, an IRB review process is currently underway to extend this survey to a \textit{large scale children survey}, where we plan to collect response data for each task from children of the ages recommended for the corresponding \nihtoolbox\ measure.

To this end, we collaborated with expert psychologists to develop child-friendly web interfaces for selected tasks and prepared them for deployment on the online developmental research platform Children Helping Science (CHS) \cite{noauthor_children_nodate}. CHS is a widely used, home-based platform through which families can participate in browser-based developmental studies run by researchers worldwide. By adapting our SAYCam-based tasks (PV, VDR and Memory) to CHS, we aim to collect performance from young children under conditions analogous to the \nihtoolbox. At the time of writing, the studies are under review and not yet live. We show two examples of our task UI design in Figures~\ref{fig:pvt-ui} and \ref{fig:vdr-ui}.

Taking PV as an example (Figure~\ref{fig:pvt-ui}), to approximate the modality of the original \nihtoolbox\ task, which relies on audio-visual interaction with spoken prompts and observed child responses, we design an \textit{audio\&video test page} to verify that instructions and target words can be delivered clearly via audio and that the child’s webcam setup is functioning for basic participation monitoring. The \textit{instruction page} provides caregiver-friendly guidance in both text and spoken form. Finally, the \textit{trial pages} present each example in a clean 2×2 grid of four large image options, paired with an audio prompt of the target word, optimizing engagement and accessibility for infants and toddlers while staying faithful to the original task format.

Following the PV setup, VDR also has an initial audio \& video test page, along with an instruction page to provide context of the experiment to the caregiver. The trial page for this task (see Figure~\ref{fig:vdr-ui}) displays the object that should be tracked, along with the video clip itself and two selectable arrows to submit an answer. Since MP4 with interactive display is not yet supported on the website, a GIF is created in its place. The beginning 5 seconds of the GIF show the first frame with a countdown, then the clip is played as normal and followed by another 5-second buffer to show that the video has ended. To help the caregiver and child understand the experiment, an interactive demo is played as the first 3 trials to showcase how each one should be properly done.\\

\section{Additional experiments \& details}
\label{app:additionalexp}

\input{author-kit-CVPR2026-v1-latex-/tables/exp_out_of_domain}
\input{author-kit-CVPR2026-v1-latex-/tables/exp_nih_results}

\subsection{Out-Of-Domain evaluation}
To test \ourmodel's capability of generalizing to unseen data domain, we further evaluate it on a set of out-of-domain (OOD) tasks that share the same structure as the in-domain benchmarks but differ in their visual domains. We consider two OOD settings: (1) \textbf{Ego4D-based tasks} use egocentric videos from the Ego4D dataset~\cite{grauman2022ego4d}, which remain first-person and naturalistic but introduce distinct environments and contexts. (2) \textbf{BabyToolbox-based tasks} correspond directly to standardized developmental psychology and clinical assessments, where the visual stimuli are abstract, non-egocentric cartoon images. The detailed test results are reflected in Table~\ref{tab:ego4d} and Table~\ref{tab:nih_toolbox_results}.\\

\begin{figure}
    \includegraphics[width=\linewidth]{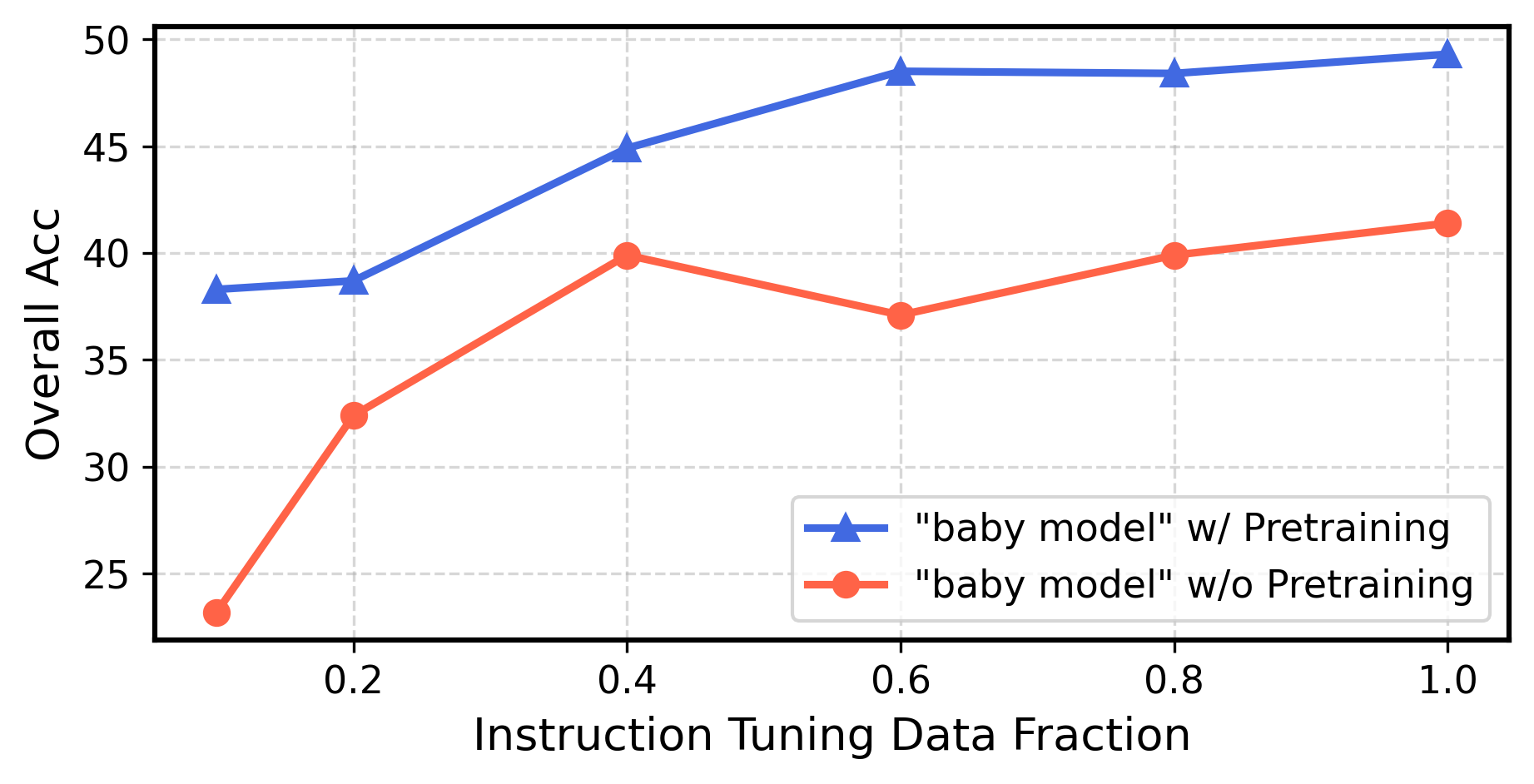}
    \caption{\ourbenchmark\ overall performance on different instruction tuning data fraction. Results are reported on epoch 1.}
    \label{fig:data_scale}
\end{figure}

\subsection{Importance of the pretraining stage}

To evaluate the contribution of the pretraining stage, we compare two variants of \ourmodel: (1) the full model trained with Stage~0--2 pretraining before instruction tuning, and (2) a randomly initialized model that skips pretraining and is trained only with Stage~3. For both variants, we fine-tune using different fractions of the instruction dataset and evaluate each model on in-domain tasks.  

As shown in Figure~\ref{fig:data_scale}, the pretrained model consistently outperforms the non-pretrained variant across all data fractions. The gap is especially pronounced when the instruction data is limited, demonstrating that pretraining provides a strong and sample-efficient initialization for downstream instruction tuning. As the instruction data fraction increases, both models improve, reflecting a clear scaling-like trend qualitatively consistent with observations in large-scale model studies~\cite{kaplan2020scalinglawsneurallanguage,hoffmann2022trainingcomputeoptimallargelanguage}. This suggests that data-dependent performance gains also exist in compact, developmentally inspired models, while pretraining remains a crucial component for achieving data-efficient learning.\\
\input{author-kit-CVPR2026-v1-latex-/tables/ablation_pretrain_dataset_threshold}

\subsection{Effect of preprocessing pretraining dataset}
\label{sec/pretrain_dataset_threshold}

To further study the effect of applying various filters to curate the pretraining dataset, we constructed two more pretraining datasets: one without any preprocessing (no filter), and the other is preprocessed by stricter filters, keeping less than half the size of the original pretraining dataset. The evaluation results of \ourmodel\ pretrained on these additional pretraining datasets are shown in Table~\ref{tab:pretrain_threshold}. \ourmodel-no filter degrades the performance a lot, which verifies the effectiveness of our pretraining dataset preprocessing pipeline; \ourmodel-strict filter shows some further improvement, suggesting potential room to refine the data preprocessing pipeline.\\

\input{author-kit-CVPR2026-v1-latex-/tables/ablation_instruction_ft}

\subsection{Instruction fine-tuning strategy ablation}
We compare two fine-tuning strategies using our instruction tuning data on \ourmodel: (1) fine-tuning a separate model for each task, and (2) jointly fine-tuning a single model on the merged instruction dataset. Table~\ref{tab:separate_vs_mix} presents the comparison.
Overall, the task-specific strategy performs slightly better than the mixed strategy. However, the performance gap is generally small across most tasks. Notably, certain tasks, such as \taskLeftRight, \taskSpatialDetails, and \taskMemory, benefit significantly from mixed fine-tuning, suggesting potential knowledge transfer or regularization effects across tasks.\\

\subsection{Synthetic caption generation}
\begin{figure}[ht]
    \centering
    \includegraphics[width=0.95\linewidth]{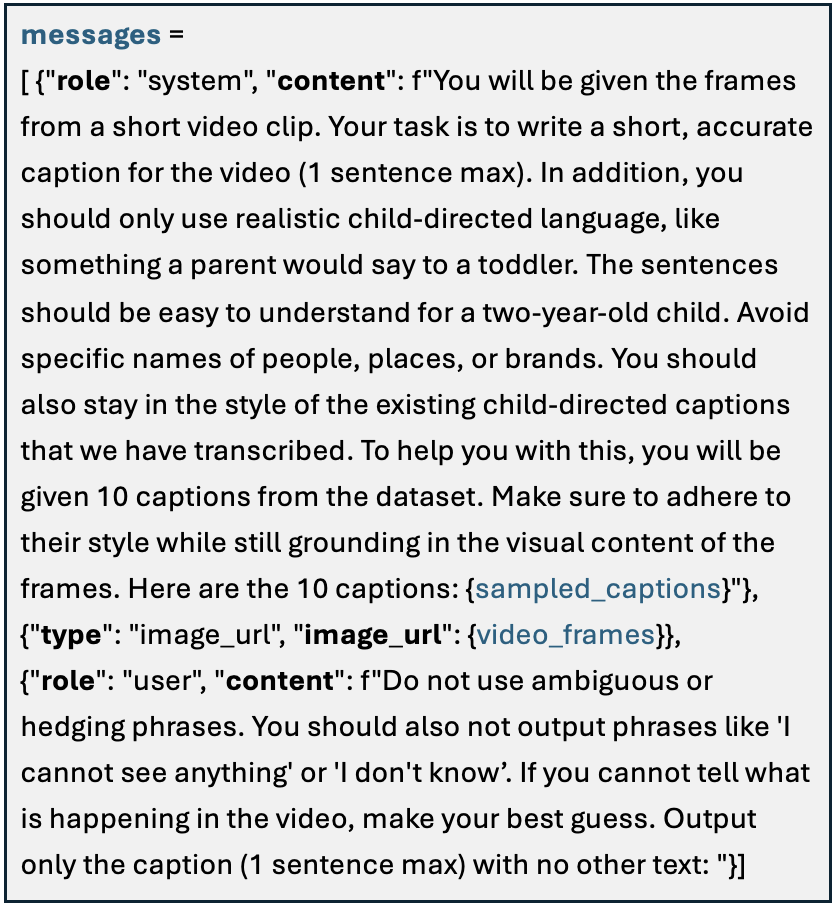}
    \caption{Full prompt for pretraining data ablation}
    \label{fig:pretrain_data_ablation}
\end{figure}

\begin{figure}[ht]
    \centering
    \includegraphics[width=0.8\linewidth]{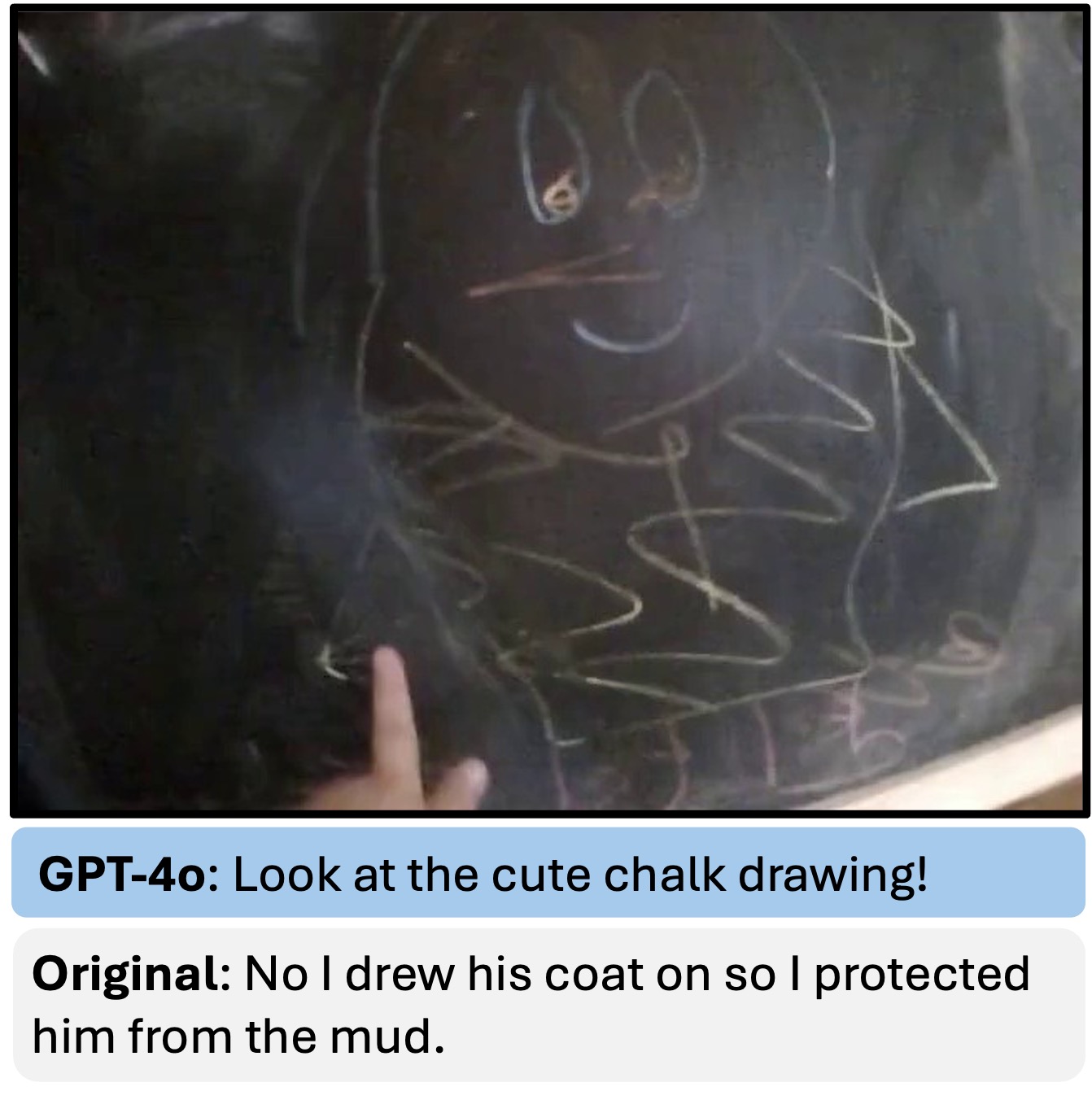}
    \caption{Example of caption generated by GPT-4o}
    \label{fig:pretrain_data_ablation_ex}
\end{figure}

We study the impact of noisy visual-alignment in the naturalistic child-directed utterances transcribed in the pretraining dataset by replacing them with video captions generated by GPT-4o. To encourage diversity in the generated captions and ensure they remain close to the style of the original dataset, we include 10 randomly sampled transcriptions in each prompt. The transcriptions are sampled from a pool of the 1,000 highest confidence transcriptions in the original dataset that contain at least one noun and more than three words. These heuristic filters help ensure that the sampled transcriptions contain stylistic information rather than simple phrases that are common in the dataset like "wow" or "let's go". The pool of 1,000 transcriptions are manually screened to remove uninformative transcriptions that passed the filtering step. The full prompt to GPT-4o is shown in Figure~\ref{fig:pretrain_data_ablation} and an example of a generated caption is shown in Figure~\ref{fig:pretrain_data_ablation_ex}.

\subsection{Prompting experiment}
\input{author-kit-CVPR2026-v1-latex-/tables/prompt_exper}
Finally, we complete a prompting experiment to show the stability of \ourbenchmark\ examples with respect to commercial models, the results of which are shown in Table~\ref{tab:prompt_exp}. We select \taskLeftRight\ and \taskCounting\ for this experiment, as we found that commercial models had the lowest and most variable performance on these. For both tasks, 100 examples are randomly selected and presented to Gemini-2.5-flash with a standard prompt, a one-shot prompt, and two variations of the standard prompt, called \textit{alternate prompt 1} and \textit{alternate prompt 2}. The standard prompt is  the one used in all other experiments, and the one shot-prompt is a prompt that includes one other example, with its correct answer, prepended to the standard prompt.

For \taskCounting, \textit{alternate prompt 1} does not give the object's name to be counted, e.g. \texttt{"<image> How many objects do you see?"}, which we see drops performance, which is intuitive because large models thrive on context, in this case the name of the object to be counted. \textit{Alternate prompt 2} gives more detail, e.g. \texttt{<image> count the flora very closely, starting from one. Keep track of which ones have already been counted and what number you've counted to thus far. Then, report how many flora you counted."}. Unsurprisingly, \textit{alternate prompt 2} does not improve performance, showing that 1) the \textit{standard} prompt was sufficient and 2) Gemini-2.5-flash has capable instruction-following capabilities. For \taskCounting, we find that a one-shot prompt does not boost performance.

For \taskLeftRight, the \textit{standard} prompt gives each image token interleaved with their answer labels, e.g. \texttt{"<image> Which of the following is the same as this? (A) <image> (B) <image> (C) <image>"}. In \textit{alternate prompt 1}, we undo this interleaving, resulting in \texttt{"<image><image><image><image> Which of the following is the same as the first one? (A) the second one, (B) the third one, or (C) the fourth one?"}. In \textit{alternate prompt 2}, we interleave even more, by giving some descriptive text before the prompt image, e.g. \texttt{"Here is an image: <image>. Which of the following is the same as it? (A) <image>, (B) <image>, or (C) <image>?"}. Intuitively we expect \textit{alternate prompt 2} to be the easiest, \textit{alternate prompt 1} to be the hardest, and the \textit{standard} prompt to fall in between. However, we find that none of these prompts elicits significantly different performance, however, the one-shot prompt \textit{significantly} boosts performance. These two findings show the robustness of Gemini-2.5-flash, and the complexity of \taskLeftRight, respectively.\\

\subsection{Effect of reasoning chain on proprietary models}
\label{app:reasoning}
Several proprietary models evaluated on \ourbenchmark\ rely on built-in reasoning chains by default, including Gemini-2.5-Flash/Pro and GPT-5. To investigate the effect of reasoning depth, we vary the \textit{thinking budget}, a hyperparameter specifying the number of tokens to use for reasoning, of Gemini-2.5-Flash from 0 to 25,000 tokens. The results reveal a striking task-dependent divergence: performance on \taskSpatialDetails, which demands fine-grained perceptual matching, monotonically increases with thinking budget, from 87.5\% to 92.1\%. In contrast, \taskWhoHasMore\ (Naturalistic), which relies on direct visual perception, sees performance decrease, dropping from 92.0\% at zero budget to 87.5\% at the maximum budget. This suggests that \ourbenchmark\ tasks differ meaningfully in the reasoning depth they require, and that extended chain-of-thought is not universally beneficial as pure perceptual tasks may be better served by direct, fast responses.
\begin{figure}[ht]
\centering
\includegraphics[width=\linewidth]{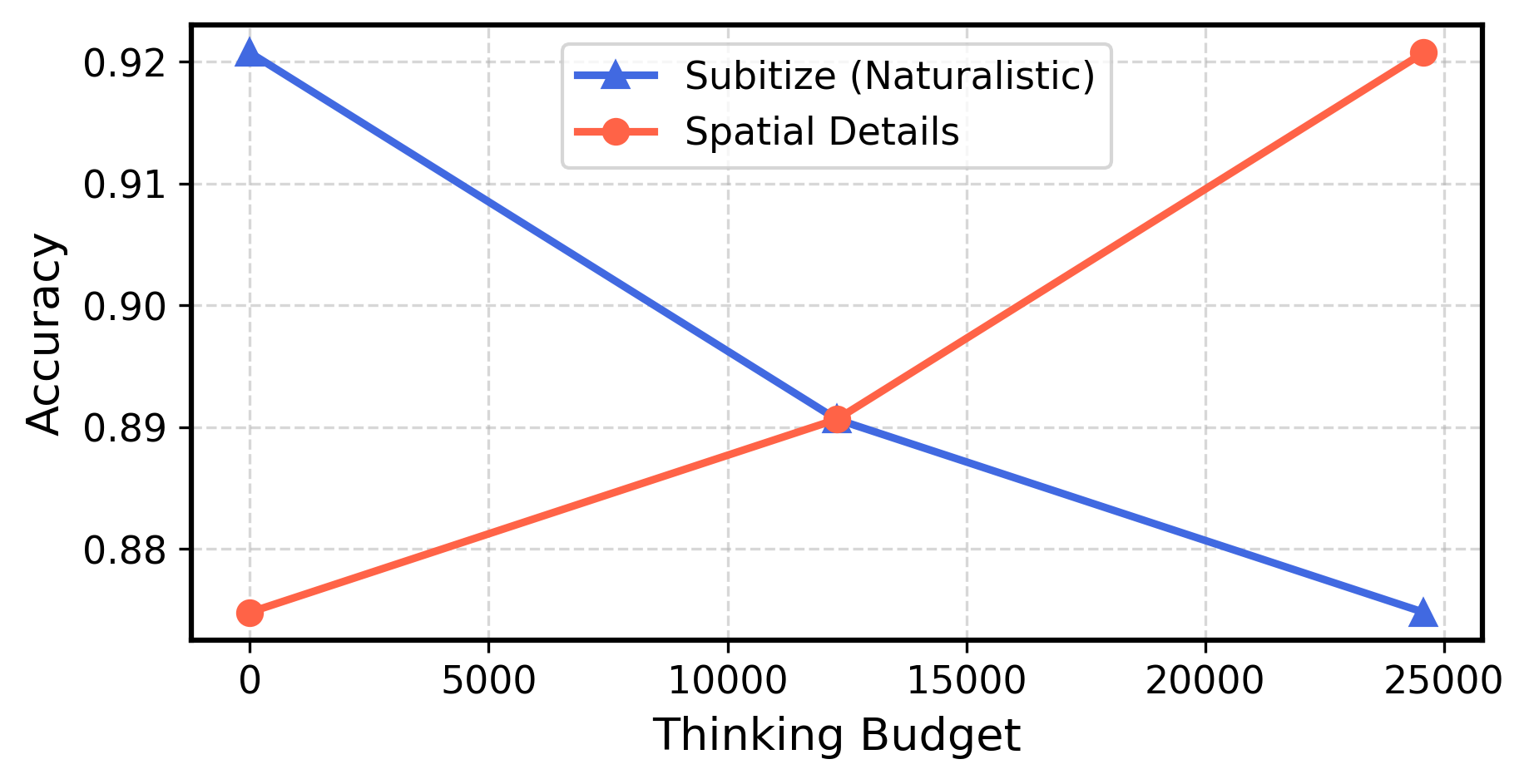}
\caption{Gemini-2.5-Flash accuracy on \ourbenchmark\ tasks across different thinking budget configurations.}
\label{fig:reasoning}
\end{figure}

\section{Ethical considerations}
\label{app:ethics}
SAYCam provides a clear license and practice guidelines for us to follow throughout this work. Whenever ambiguities arose, we consulted directly with the SAYCam authors, who graciously served as consultants to the project through multiple meetings. We have confirmed that the experiments and data collection procedures described in this paper do not require IRB approval. However, the large-scale children survey on \ourbenchmark\ that we plan as future work does require IRB, and we are actively working with a psychologist and the IRB office to obtain the necessary approvals.

%% file: author-kit-CVPR2026-v1-latex-/tables/training_paradigm.tex
\begin{table*}
\footnotesize
\centering
\caption{\textbf{Training stage specification of \ourmodel}. Note that for stage 3, different modules have different learning rate, as mentioned in Section~\ref{sec:training_paradigm}.}
\vspace{-10pt}
\label{tab:training_paradigm}

\resizebox{\textwidth}{!}{
\begin{tabular}{lccccccc}
\toprule

{\textbf{Stage}} & \textbf{Trained modules} & \textbf{Frozen modules} & \textbf{Dataset} & \textbf{Loss} & \textbf{Learning rate} & \textbf{Epoch} & \textbf{Global batch size}\\
\midrule

0-language & Language backbone & N/A & 283k utterance only & Autoregressive & 2e-4 & 10 & 16\\
\midrule
0-vision & Vision backbone & N/A & 1085k image only & DINOv2 & 1e-4 & 100 & 64\\
\midrule
1 & MLP connector & \makecell{Language backbone\\ + vision backbone} & 768k image-utterance & Autoregressive & 3e-3 & 5 & 128\\
\midrule
2 & \makecell{MLP connector \\+ language backbone} & Vision backbone & \makecell{768k image-utterance\\ + 181k video-utterance \\+ 63k multi-turn} & Autoregressive & 2e-4 & 5 & 128\\
\midrule
3 & \makecell{MLP connector \\+ language backbone \\+ vision backbone} & None & \makecell{113k instruction finetune} & Autoregressive & 5e-4 $\mid$ 1e-4 & 5 & 128\\

\bottomrule
\end{tabular}
}

\vspace{-10pt}
\end{table*}

%% file: author-kit-CVPR2026-v1-latex-/tables/comparison.tex
\begin{table}[t]
\footnotesize
\centering
\caption{\textbf{Comparison between SAYCam and Ego4d}. Object Size is reported in terms of the average \% of the frame's area filled.}
\vspace{-10pt}
\label{tab:comparison}

\begin{tabular}{lccc}
\toprule

{\textbf{Data Source}} & \textbf{participants} & \textbf{Number of Pixels} & \textbf{Object Size}  \\
\midrule

SAYCam &infants & 307k (fixed) &57\% \\

Ego4d &adults &  over 2M (average)&4\% \\

\bottomrule
\end{tabular}

\vspace{-10pt}
\end{table}


%% file: author-kit-CVPR2026-v1-latex-/tables/exp_out_of_domain.tex
\begin{table*}
\centering
\caption{\textbf{Performance comparison across models on \ourbenchmark\ out-of-domain tasks (Ego4D).}
Different background colors denote different model families. 
We report accuracy (\%) for all tasks. }
\vspace{-10pt}
\label{tab:main_result_ood}
\resizebox{\textwidth}{!}{
\begin{tabular}{lccccccccccccc}
\toprule
\multirow{2}{*}{\textbf{Model}} &
\multirow{2}{*}{\textbf{Overall}} &
\multirow{2}{*}{\textbf{Count}} &
\multirow{2}{*}{\textbf{LeftRight}} &
\multirow{2}{*}{\textbf{Spatial}} &
\multirow{2}{*}{\textbf{PV}} &
\multirow{2}{*}{\textbf{Memory}} &
\multirow{2}{*}{\textbf{Localization}} &
\multicolumn{3}{c}{\textbf{Visual Delay Response}} &
\multicolumn{2}{c}{\textbf{Who Has More}} \\
\cmidrule(lr){9-11} \cmidrule(lr){12-13}
 &  &  &  &  &  &  &  & \textbf{binary} & \textbf{multi-exact} & \textbf{multi-adjacent} & \textbf{synthetic} & \textbf{naturalistic} \\
\midrule

\rowcolor{groupC}
\multicolumn{13}{l}{\textbf{Upper Bound}} \\
\rowcolor{groupC}
{\tt Human performance}  &  {\tt 93.5} & {\tt 96.4} & {\tt 98.2} & {\tt 96.4} & {\tt 96.4} & {\tt 98.8} & {\tt 90.9} & {\tt 100} & {\tt 58.2} & {\tt 100} & {\tt 100} & {\tt 92.7} \\[5pt]

\midrule
\rowcolor{groupA}
\multicolumn{13}{l}{\textbf{Proprietary models}} \\
\rowcolor{groupA}
GPT-4o  &  67.6  &  62.1 &  45.1 & 94.7 & 85.3 &  \textbf{100} & 80.4 & 45.5 & 13.2 & 48.3 & 84.3 & 84.6 \\
\rowcolor{groupA}
GPT-5   &  86.7 & 77.5  & \textbf{88.0} & \textbf{96.8} & 91.9 &  \textbf{100} & 88.7 & \textbf{94.4} &  \textbf{50.3}  & 82.6 & 94.6 & 88.5 \\
\rowcolor{groupA}
Gemini-2.5-flash  &  77.7 &  72.9  & 49.6 & 86.7 & \textbf{92.5} &  99.2 & 88.4 & 80.6 & 37.1 & 70.2 & \textbf{97.8} & 80.1 \\
\rowcolor{groupA}
Gemini-2.5-pro  &  \textbf{88.2} & \textbf{81.9}  & \textbf{88.0}  & 94.8 & 91.9 & \textbf{100}  & \textbf{90.2} & 91.3 & \textbf{50.3} & \textbf{87.9} & 96.5 & \textbf{97.8} \\[5pt]
\midrule
\rowcolor{groupB}
\multicolumn{13}{l}{\textbf{Open-source models  - Similar Size as Ours}} \\
\rowcolor{groupB}
LLaVA-OneVision-0.5B   &  39.4 &  \underline{43.9} &  32.6  & 33.3  & 27.7  & 22.6  & 21.6 &  73.0 & 15.2  & 67.7  &  46.8 & 49.4 \\


\rowcolor{groupB}
InternVL3.5-1B   & 43.7  &  34.7  & 34.0 & 34.1  &  33.8 & 24.9  & 60.7  & 73.9 &  16.9 &  \underline{68.5} & 49.0 & 49.9 \\

\rowcolor{groupB}
Qwen2.5-VL-3B   &  48.1 & 35.7  &  32.6  & 44.1  &  \underline{41.9} &  25.7  & \underline{86.7} & \underline{79.8}  &  \underline{28.9} &  51.1 & 50.2  & \underline{53.4} \\[5pt]


\rowcolor{groupD}
\multicolumn{13}{l}{\textbf{Ours}} \\
\rowcolor{groupD}
\ourmodel    & \underline{48.8}  &  42.1 &  \underline{82.2} &  \underline{47.1}  &  23.4  & \underline{72.4}  & 27.7  &  39.9 & 22.8 & 36.2 & \underline{92.5}  & 50.9 \\
[5pt]
\midrule
\rowcolor{groupE}
\multicolumn{13}{l}{\textbf{Lower Bound}} \\
\rowcolor{groupE}
{\tt Random guess} & {\tt 31.8} & {\tt 8.33} & {\tt 33.3} & {\tt 33.3} & {\tt 25.0} & {\tt 25.0} & {\tt 25.0} & {\tt 50.0} & {\tt 12.5} & {\tt 37.5} & {\tt 50.0} & {\tt 50.0} \\

\bottomrule
\end{tabular}
\label{tab:ego4d}
}
\end{table*}

%% file: author-kit-CVPR2026-v1-latex-/tables/exp_nih_results.tex
\begin{table}[t]
\footnotesize
\centering
\caption{\textbf{Performance on NIH Baby Toolbox out-of-domain tasks.} We report the \#correct/\#total for all tasks.}
\vspace{-10pt}
\label{tab:nih_toolbox_results}

\resizebox{0.5\textwidth}{!}{
\begin{tabular}{lccc}
\toprule

{\textbf{Model}} & \textbf{Who Has More} & \textbf{Count} & \textbf{Mullen Visual Reception}  \\
\midrule

\ourmodel & 13/24 & 2/6 & 3/12 \\

\bottomrule
\end{tabular}
}
\vspace{-10pt}
\end{table}


%% file: author-kit-CVPR2026-v1-latex-/tables/ablation_pretrain_dataset_threshold.tex
\begin{table*}[ht]
\centering
\vspace{-10pt}
\caption{\textbf{Ablation of pretraining data filtering threshold on \ourbenchmark.} \ourmodel\ is the model pretrained on the regular pretraining dataset described in the main paper, \ourmodel-no filter and \ourmodel-strict filter are models pretrained on different pretraining datasets described in Section~ \ref{sec/pretrain_dataset_threshold}.}
\vspace{-10pt}
\label{tab:pretrain_threshold}
\resizebox{\textwidth}{!}{
\begin{tabular}{lc!{\vrule}ccccccccccc}
\toprule
\multirow{2}{*}{\textbf{Model}} &
\multirow{2}{*}{\textbf{Overall}} &
\multirow{2}{*}{\textbf{Count}} &
\multirow{2}{*}{\textbf{LeftRight}} &
\multirow{2}{*}{\textbf{Spatial}} &
\multirow{2}{*}{\textbf{PV}} &
\multirow{2}{*}{\textbf{Memory}} &
\multirow{2}{*}{\textbf{Localization}} &
\multicolumn{3}{c}{\textbf{Visual Delay Response}} &
\multicolumn{2}{c}{\textbf{Who Has More}} \\
\cline{9-11} \cline{12-13}
 &  &  &  &  &  &  &  & \textbf{binary} & \textbf{multi-exact} & \textbf{multi-adjacent} & \textbf{synthetic} & \textbf{naturalistic} \\
\midrule

\ourmodel  &  63.9 &  47.3 & 96.4  & 92.8  &  32.4 & 90.8  & 37.8  & 54.6  & 38.1  &  \textbf{52.5} & 99.7  &  60.5 \\
\ourmodel-no filter  &  56.0  &  46.5 &  41.8 & \textbf{95.9} &  32.6 &  63.1  &  38.2  &  54.2  & 36.1  &  49.5 & \textbf{99.8}  &  58.6 \\
\ourmodel-strict filter  &  \textbf{67.3} &  \textbf{53.2} & \textbf{99.7}  & 95.8  &  \textbf{33.3} & \textbf{94.0}  & \textbf{44.7}  & \textbf{58.5}  & \textbf{40.2}  &  52.0 & \textbf{99.8}  &  \textbf{68.8} \\
\bottomrule
\end{tabular}
}
\end{table*}

%% file: author-kit-CVPR2026-v1-latex-/tables/ablation_instruction_ft.tex
\begin{table*}
\centering
\caption{\textbf{Two supervised fine-tuning strategies}.
\ourmodel-separate denotes models fine-tuned on each task's instruction dataset separately, and \ourmodel-mixed is a single model fine-tuned on the mixed instruction set.
}
\vspace{-10pt}
\label{tab:separate_vs_mix}
\resizebox{\textwidth}{!}{
\begin{tabular}{lc!{\vrule}ccccccccccc}
\toprule
\multirow{2}{*}{\textbf{Model}} &
\multirow{2}{*}{\textbf{Overall}} &
\multirow{2}{*}{\textbf{Count}} &
\multirow{2}{*}{\textbf{LeftRight}} &
\multirow{2}{*}{\textbf{Spatial}} &
\multirow{2}{*}{\textbf{PV}} &
\multirow{2}{*}{\textbf{Memory}} &
\multirow{2}{*}{\textbf{Localization}} &
\multicolumn{3}{c}{\textbf{Visual Delay Response}} &
\multicolumn{2}{c}{\textbf{Who Has More}} \\
\cline{9-11} \cline{12-13}
 &  &  &  &  &  &  &  & \textbf{binary} & \textbf{multi-exact} & \textbf{multi-adjacent} & \textbf{synthetic} & \textbf{naturalistic} \\
\midrule

\ourmodel-separate  &  56.9  &  45.2 &  54.9 & 79.8 &  \textbf{32.5} &  73.6  &  36.3  &  \textbf{55.7}  & 37.0  &  49.9  &  98.6 & \textbf{62.1} \\
\ourmodel-mixed  &  \textbf{63.9} &  \textbf{47.3} & \textbf{96.4}  & \textbf{92.8}  &  32.4 & \textbf{90.8}  & \textbf{37.8}  & 54.6  & \textbf{38.1}  &  \textbf{52.5} & \textbf{99.7}  &  60.5 \\
\bottomrule
\end{tabular}
}
\end{table*}

%% file: author-kit-CVPR2026-v1-latex-/tables/prompt_exper.tex
\begin{table}
\label{table:prompts}
\centering
\caption{\textbf{Comparison between Gemini-2.5-flash performance with different prompting strategies}
}
\vspace{-10pt}

\begin{tabular}{lcc}
\toprule
\multirow{2}{*}{\textbf{Prompt Type}} &

\multirow{2}{*}{\textbf{Count}} &
\multirow{2}{*}{\textbf{LeftRight}} \\
\vspace{5pt}\\

\midrule

standard   &  69 &  55  \\
one-shot   &  66 &   82  \\
alternate prompt 1  & 55 &54     \\
alternate prompt 2 & 67 & 56 \\
\bottomrule
\end{tabular}
\label{tab:prompt_exp}
\end{table}


